\documentclass[sigconf]{acmart}
\usepackage{algorithm}
\usepackage{algpseudocode}
\usepackage{multirow}
\usepackage{booktabs}
\usepackage{siunitx}
\usepackage{tikz}
\usepackage[dvipsnames]{xcolor}
\usetikzlibrary{positioning}
\usepackage{calc}
\usepackage{array}
\usepackage{makecell}
\usepackage{stfloats}

\acmDOI{}
\acmISBN{}
\acmConference[]{}{}{}{}   % 注意有 []，把 short name 也清空
\acmBooktitle{}            % 避免 booktitle 参与拼接
\acmYear{}                 % 可选：避免年份参与拼接

\setcopyright{none}

\renewcommand\footnotetextcopyrightpermission[1]{}

\AtBeginDocument{%
  }

\begin{document}

\title{EXG: Self-Evolving Agents with Experience Graphs}

\author{Yuxin Jin}
\affiliation{%
  \institution{University of Technology Sydney}
  \city{Sydney}
  % \state{New South Wales}
  \country{Australia}
}
\email{yuxin.jin-1@student.uts.edu.au}

\author{Siyuan Zhang}
\affiliation{%
  \institution{University of Technology Sydney}
  \city{Sydney}
  % \state{New South Wales}
  \country{Australia}
}
\email{Siyuan.Zhang@uts.edu.au}

\author{Hanchen Wang}
% \authornote{Corresponding author.}
\affiliation{%
  \institution{University of Technology Sydney}
  \city{Sydney}
  % \state{New South Wales}
  \country{Australia}
}
\email{hanchen.wang@uts.edu.au}

\author{Lu Qin}
\affiliation{%
  \institution{University of Technology Sydney}
  \city{Sydney}
  % \state{New South Wales}
  \country{Australia}
}
\email{Lu.Qin@uts.edu.au}

\author{Ying Zhang}
\affiliation{%
  \institution{University of Technology Sydney}
  \city{Sydney}
  % \state{New South Wales}
  \country{Australia}
}
\email{Ying.Zhang@uts.edu.au}

\author{Wenjie Zhang}
\affiliation{%
  \institution{The University of New South Wales}
  \city{Sydney}
  % \state{New South Wales}
  \country{Australia}
}
\email{wenjie.zhang@unsw.edu.au}


\begin{abstract}
  Large language model (LLM)–based agents have demonstrated strong capabilities in complex reasoning and problem solving through multi-step interactions, yet most deployed agents remain behaviorally static, with knowledge acquired during execution rarely translating into systematic improvement over time. In response, a growing line of work on self-evolving agents explores how agents can improve through experience during deployment, but most existing approaches either rely on ad hoc reflection limited to single-task correction or adopt unstructured memory that accumulates fragmented experience with delayed usability. To address this limitation, we introduce EXG, an experience graph framework for self-evolving agents that explicitly organizes accumulated successes and failures into a structured, relational representation. EXG is the first experience graph designed for self-evolving agents, supporting both online, real-time graph growth during execution for immediate cross-task experience reuse, and offline reuse of a consolidated experience graph as an external memory module. This design also enables EXG to serve as a plug-and-play component for existing self-evolving agents, organizing prior experience into a unified experience graph and improving both solution quality and resource efficiency as deployment progresses. Extensive experiments across code generation and reasoning benchmarks show that EXG attains more favorable performance–efficiency trade-offs than reflection- and memory-based baselines in both online and offline evaluations. Our results suggest that structuring experience as a graph provides a principled foundation for scalable and transferable self-evolving agent behavior.
\end{abstract}

\maketitle

\section{Introduction}

\begin{figure*}[t]
    \centering
    \includegraphics[
        width=0.8\textwidth,
        % height=8cm
        % trim={0 0.05cm 0 0},
        % clip
    ]{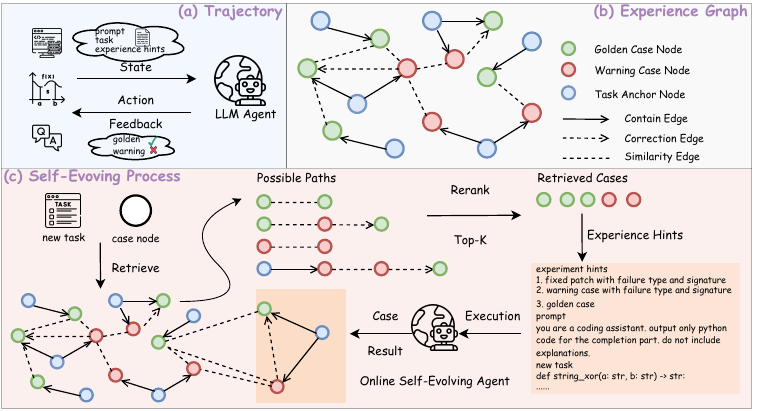}
    \caption{Overview of the self-evolving experience graph. (a) Trajectory produces structured cases from agent interactions. (b) Experience graph organizes cases into a relational structure. (c) Self-evolving process reuses retrieved experience across tasks and continuously incorporates resulting experience back into the graph.}
    \label{fig1}
\end{figure*}

Recent advances in large language models (LLMs) have enabled agents to function as interactive problem solvers capable of sustaining long sequences of reasoning and action, making them suitable for tasks that involve extended interaction and non-trivial planning \cite{10.5555/3600270.3602070, yao2023reactsynergizingreasoningacting, wang2023selfconsistencyimproveschainthought}. These advances have fueled rapid progress in autonomous agents across domains such as code generation and multi-hop question answering \cite{chen2021evaluatinglargelanguagemodels,yang-etal-2018-hotpotqa}. Despite this progress, most deployed agents remain fundamentally static: each task is treated largely in isolation, and knowledge acquired during previous attempts rarely translates into systematic improvement on future tasks \cite{10.1145/3586183.3606763, packer2024memgptllmsoperatingsystems, wang2023voyageropenendedembodiedagent}. As a result, agents often repeat similar mistakes, fail to consolidate past successes, and incur growing computational costs as deployment continues \cite{NEURIPS2023_91edff07, NEURIPS2023_1b44b878, ICLR2024_3339f19c}. 

To overcome the static nature of deployed agents, recent research has begun to explore self-evolving agents, which seek to improve future task performance by leveraging experience acquired through an agent’s own interactions. Rather than relying on parameter updates \cite{wu2025evolverselfevolvingllmagents,zhang2025agentlearningearlyexperience}, a prominent line of work studies non-parametric self-evolution in two settings, with online self-evolution \cite{NEURIPS2023_1b44b878, lin2025seagentselfevolutiontrajectoryoptimization} generating and consuming experience within individual tasks during inference, and offline self-evolution \cite{liu-etal-2025-contextual, 10.1609/aaai.v38i17.29936, ouyang2025reasoningbankscalingagentselfevolving, cao2025remembermerefineme} accumulating experience across tasks and reusing it over time to guide subsequent reasoning. However, existing approaches along both directions face more fundamental limitations. Online self-evolving methods typically confine experience to a single task, lacking mechanisms to persist and reuse experience across tasks during inference. 
Offline self-evolving methods, while able to consolidate experience collected online and reuse it at test time, usually depend on time-consuming and complex post-processing.
The experience generated during online interaction is often not directly usable. 
More broadly, self-evolving agents lack a unified framework for organizing experience that simultaneously supports online cross-task reuse and serves as an effective offline memory module for new tasks. 

Motivated by these limitations, the design of experience graphs (EXG) is guided by three corresponding principles. Rather than confining experience to individual tasks, EXG externalizes interaction outcomes into a persistent experience graph, allowing knowledge acquired during one task to be immediately reused by subsequent tasks and enabling online self-evolution beyond task boundaries. At the same time, by maintaining a unified graph representation throughout deployment, EXG avoids the costly and ad hoc post-processing typically required by offline methods, making online experience directly reusable as an external memory at test time. EXG further provides a single, graph-centric abstraction for organizing experience that bridges online and offline self-evolution, while also establishing a plug-and-play foundation that allows existing self-evolving agents to incorporate structured experience without modifying their underlying architectures.

We propose EXG as a self-evolution mechanism that incrementally structures interaction experience during agent execution. Concretely, EXG abstracts each attempt within a task into a structured case and inserts it as a node in a growing experience graph, where edges encode relational signals such as task association, semantic similarity, and error correction. As the agent operates, newly generated experience is immediately integrated into this shared graph rather than remaining local to the current task. For each new task, EXG retrieves and reranks relevant cases from the graph to construct experience hints that guide reasoning, after which the resulting interaction is added back to the graph. This online self-evolving loop enables experience to accumulate and compound across tasks while remaining directly usable at inference time. 
Because a unified graph representation is maintained throughout deployment, the experience graph can be reused offline as an external memory module without additional consolidation or retraining, enabling EXG to bridge online and offline self-evolution within a single plug-and-play architecture. 
Figure~\ref{fig1} shows the construction of the experience graph and how experience is reused. 

Our contributions can be summarized as follows:
\begin{itemize}
    \item \textbf{Experience Graph for Self-Evolving Agents.} We introduce EXG, the first graph-based mechanism for self-evolving agents that explicitly encodes experience into a structured representation.

    \item \textbf{Unified Online and Offline Self-Evolution.} EXG bridges online self-evolution and cross-task experience reuse within a single representation by incrementally constructing the experience graph during execution and directly reusing the resulting graph as an external memory in offline settings.

    \item \textbf{Plug-and-Play Experience Organization.} By operating solely at inference time and remaining external to the agent’s internal reasoning mechanism, EXG serves as a plug-and-play module that can be seamlessly integrated into existing self-evolving agents, improving their ability to exploit accumulated experience without modifying underlying models.

    \item \textbf{Higher Accuracy at Lower Computational Cost.} We demonstrate through extensive experiments on reasoning and code generation benchmarks that structuring experience with EXG yields higher pass@1 exceeding \textbf{150\%} and pass@2 approaching \textbf{30\%} in the online setting and comparable performance in the offline setting, while substantially reducing interaction cost, with up to \textbf{45.7\%} fewer LLM calls and up to \textbf{30.5\%} lower LLM inference latency.

\end{itemize}
 
\section{Experience Graph Design}
\subsection{Trajectories and Case Abstraction}
\label{case}
An agent’s interaction with a task is naturally represented as a trajectory consisting of alternating reasoning, action, and feedback signals produced during execution, as illustrated in Figure~\ref{fig1} (a). Formally, for a given task \( \tau \), we denote an interaction trajectory as
\begin{equation}
\mathcal{T}_\tau = \{(s_t, a_t, o_t)\}_{t=1}^{T},
\end{equation}
where \( s_t \) represents the agent’s internal state or prompt context at step \( t \), \( a_t \) denotes the action taken by the agent, and \( o_t \) is the observed outcome or feedback from the environment, such as execution results or correctness signals. The trajectory terminates after a finite number of steps when a solution is produced or a failure condition is reached. While this trajectory-level representation captures the full interaction process, it remains unstructured and task-local, making it unsuitable for direct reuse across tasks.

To enable structured reuse, EXG abstracts each completed attempt within a trajectory into a \emph{case}, which serves as the atomic unit of experience in the graph. 
Specifically, given a trajectory \( \mathcal{T}^{(k)}_\tau \) corresponding to the \( k \)-th attempt on task \( \tau \), we construct a case
\begin{equation}
c_{\tau}^{(k)} =
\big(
\tau,\;
x_\tau,\;
y_{\tau}^{(k)},\;
r_{\tau}^{(k)},\;
\sigma_{\tau}^{(k)}
\big),
\end{equation}
where \( x_\tau \) denotes the task input, \( y_{\tau}^{(k)} \) is the agent’s final output for this attempt, \( r_{\tau}^{(k)} \in \{0,1\} \) indicates whether the attempt is successful, and \( \sigma_{\tau}^{(k)} \) summarizes salient execution signals extracted from the trajectory, such as error messages, failure types, or corrective feedback. 
Successful cases (\( r_{\tau}^{(k)} = 1 \)) are treated as \emph{golden cases}, while unsuccessful ones (\( r_{\tau}^{(k)} = 0 \)) are treated as \emph{warning cases}. 
By collapsing a full interaction trajectory into a compact structured representation, cases provide a reusable abstraction that can be incrementally organized within the experience graph.

\subsection{Experience Graph Construction}
Based on the above case abstraction, EXG organizes accumulated experience into a relational experience graph, as illustrated in Figure~\ref{fig1} (b). 
Formally, the experience graph is defined as
\begin{equation}
\mathcal{G} = (\mathcal{V}, \mathcal{E}),
\end{equation}
where the vertex set \( \mathcal{V} \) consists of heterogeneous node types representing different levels of experience abstraction, and the edge set \( \mathcal{E} \) encodes typed relations among them. 

\textit{Nodes.}
The node set \( \mathcal{V} \) includes two primary types:

\textbf{Case nodes.} Each case node corresponds to a case \( c_{\tau}^{(k)} \) constructed from an interaction trajectory, as defined in Section~\ref{case}. 
Case nodes, including \textit{golden} and \textit{warning} case nodes, serve as the atomic units of experience in the graph.

\textbf{Task anchor nodes.} For each task \( \tau \), a task anchor node \( a_\tau \) is introduced to group all cases associated with the same task, providing a task-level entry point for organizing and retrieving experience.

\textit{Edges.}
Edges in the experience graph are typed and directed, capturing distinct semantic relations between nodes. We consider the following edge types:

\textbf{Contain edges.} A directed edge \( (a_\tau \rightarrow c_{\tau}^{(k)}) \) indicates that case \( c_{\tau}^{(k)} \) is an attempt associated with task \( \tau \). These edges establish the hierarchical structure between task anchors and case nodes.

\textbf{Similarity edges.} An undirected edge \( (c_i - c_j) \) labeled as \emph{similar\_to} encodes semantic similarity between two cases. Similarity is computed based on representations derived from case attributes such as task inputs, prompts, or extracted signatures, enabling the graph to expose reusable patterns across different tasks.

\textbf{Correction edges.} For multiple attempts on the same task, a directed edge \( (c_{\tau}^{(k)} \rightarrow c_{\tau}^{(k')}) \) labeled as \emph{fixed\_by} indicates that the case $c_{\tau}^{(k')}$ corrects or resolves the failure observed in the earlier case $c_{\tau}^{(k)}$. These edges explicitly capture error–repair relationships within a task.

\subsection{Experience Retrieval}
Algorithm~\ref{alg:experience_retrieval} summarizes the retrieval procedure. Given an experience graph \( \mathcal{G}=(\mathcal{V},\mathcal{E}) \), the retrieval process constructs a bounded pool of candidate cases to support the current attempt, corresponding to the \textit{Retrieve} stage in Figure~\ref{fig1} (c).
The retrieval process is conditioned on the \emph{provisional case} \(c_q\), a partially instantiated case that contains the task input and contextual information while its output and outcome fields remain undefined.
Based on this provisional case, retrieval integrates three complementary sources, namely task-local context from \textit{task anchor cases}, semantic neighborhoods induced by \textit{similarity} edges, and corrective traces captured by \textit{correction} edges.

\textit{Task-anchor cases.}
Let \(a_{\tau(q)}\) denote the task anchor associated with \(c_q\). The task-local set is
\begin{equation}
\mathcal{C}_{\text{task}} \;=\; \{\, c \in \mathcal{V} \mid (a_{\tau(q)} \rightarrow c)\in \mathcal{E}_{\text{contain}} \,\}.
\end{equation}

\textit{Semantic seeds and one-hop expansion.}
We form semantic seeds from two channels. The query-side seed set \(\mathcal{S}_{\text{query}}\subseteq \mathcal{V}\) contains cases semantically related to \(c_q\) (e.g., via an embedding index). 
In addition, we select a subset of task-anchor cases as seeds for bridging, $
\mathcal{A}_{\text{seed}} \;=\; \mathrm{Select}(\mathcal{C}_{\text{task}})$, where \(\mathrm{Select}(\cdot)\) prioritizes warning cases when available. Bridge seeds are then obtained by traversing \textit{similar\_to} relations from \(\mathcal{A}_{\text{seed}}\),
\begin{equation}
\mathcal{S}_{\text{bridge}} \;=\; \{\, c' \in \mathcal{V} \mid \exists c \in \mathcal{A}_{\text{seed}},\; (c \rightarrow c')\in \mathcal{E}_{\text{sim}} \,\}.
\end{equation}
Let \(\mathcal{S}=\mathcal{S}_{\text{query}}\cup \mathcal{S}_{\text{bridge}}\). A bounded semantic neighborhood is collected by one-hop expansion along \textit{similarity} edges
% \textit{similar\_to} edges,
\begin{equation}
\mathcal{C}_{\text{sim}} \;=\; \{\, c' \in \mathcal{V} \mid \exists c \in \mathcal{S},\; (c - c')\in \mathcal{E}_{\text{sim}} \,\}.
\end{equation}

\textit{Corrective traces.}
To incorporate explicit error-repair relations, we follow \textit{correction} edges from semantically expanded cases and add their corrected counterparts,
\begin{equation}
\mathcal{C}_{\text{fix}} \;=\; \{\, c' \in \mathcal{V} \mid \exists c \in \mathcal{C}_{\text{sim}},\; (c \rightarrow c')\in \mathcal{E}_{\text{fix}} \,\}.
\end{equation}

\textit{Final candidate pool.}
The final retrieval pool aggregates diverse and complementary signals by combining anchor cases from the current task, semantically similar cases retrieved via seed-based expansion, and associated corrective evidence, followed by deduplication and a global cap:
\begin{equation}
\mathcal{C} \;=\; \mathrm{Cap}\!\left(\mathrm{Deduplicate}\!\left(\mathcal{C}_{\text{task}} \cup \mathcal{C}_{\text{sim}} \cup \mathcal{C}_{\text{fix}}\right)\right).
\end{equation}
This design increases the likelihood that the retrieved pool offers rich semantic context and actionable experience, enabling more effective guidance for the current case.

\begin{algorithm}[t]
\caption{Experience Retrieval}
\label{alg:experience_retrieval}
\begin{algorithmic}[1]
\Require Provisional case \(c_q\), experience graph \(\mathcal{G}\)
\Ensure Candidate case set \(\mathcal{C}\)
\State \(\mathcal{C}_{\text{task}} \leftarrow\) cases attached to the task anchor of \(c_q\)
\State \(\mathcal{A}_{\text{seed}} \leftarrow \mathrm{Select}(\mathcal{C}_{\text{task}})\)
\State \(\mathcal{S}_{\text{query}} \leftarrow\) cases semantically related to \(c_q\) \Comment{query-side seeds}
\State \(\mathcal{S}_{\text{bridge}} \leftarrow\) cases reached by traversing \textit{similar\_to} edges from \(\mathcal{A}_{\text{seed}}\) \Comment{task-side bridge seeds}
\State \(\mathcal{S} \leftarrow \mathcal{S}_{\text{query}} \cup \mathcal{S}_{\text{bridge}}\)
\State \(\mathcal{C}_{\text{sim}} \leftarrow\) one-hop expansion from \(\mathcal{S}\) along \textit{similar\_to} edges
\State \(\mathcal{C}_{\text{fix}} \leftarrow \emptyset\)
\ForAll{\(c \in \mathcal{C}_{\text{sim}}\)}
    \If{\(c\) has an outgoing \textit{correction} edge}
        \State add the destination case to \(\mathcal{C}_{\text{fix}}\)
    \EndIf
\EndFor
\State \(\mathcal{C} \leftarrow \mathrm{Deduplicate}(\mathcal{C}_{\text{task}} \cup \mathcal{C}_{\text{sim}} \cup \mathcal{C}_{\text{fix}})\)
\State \Return \(\mathrm{Cap}(\mathcal{C})\)
\end{algorithmic}
\end{algorithm}

\subsection{Experience Reranking}
Given the candidate pool \( \mathcal{C} \) returned by retrieval, reranking orders cases by a relevance score that combines prompt-level similarity with failure-aware signals when available, and further incorporates structural proximity via one-hop propagation over \textit{similarity} edges.

\textit{Case similarity.}
For each case \(c\), we derive two embeddings, a prompt embedding \( \mathbf{e}_p(c) \) from its prompt content and a failure embedding \( \mathbf{e}_f(c) \) from its failure-related text. We also define an indicator \( h(c)\in\{0,1\} \) that denotes whether \(c\) contains failure information. 
The similarity between cases \(c_i\) and \(c_j\) is defined as
\begin{equation}
s(c_i,c_j)
= \alpha \langle \mathbf{e}_p(c_i), \mathbf{e}_p(c_j)\rangle +
(1-\alpha)h(c_i)h(c_j)\langle \mathbf{e}_f(c_i), \mathbf{e}_f(c_j)\rangle,
\label{eq:case_similarity}
\end{equation}
where \( \alpha\in[0,1] \) controls the trade-off between prompt semantics and failure-aware similarity. The gate \(h(c_i)h(c_j)\) ensures that the failure term contributes only when both cases carry failure signals.

\textit{Seed initialization.}
Reranking is initialized from a seed set \( \mathcal{S} \) consisting of (i) query-side seeds semantically matched to the provisional case \(c_q\) and (ii) bridge seeds reached from warning-prioritized anchor-associated cases during retrieval. Each seed \(c\in\mathcal{S}\) is assigned an initial relevance \( \rho_0(c) \) reflecting its matching strength.

\textit{One-hop relevance propagation.}
To incorporate structural proximity, we propagate seed relevance through one-hop \textit{similarity} edges. 
For candidate case \(c \in \mathcal{C}\), we compute its final relevance as
\begin{equation}
\rho(c)=\max\Big(\rho_0(c),\ \max_{u\in \mathcal{S}}\big[\rho_0(u) + w(u,c)\big]\Big),
\label{eq:score_propagation}
\end{equation}
where \( w(u,c) \) denotes the weight of the \textit{similarity} edge from \(u\) to \(c\) in \( \mathcal{E} \). 
Intuitively, a case receives a higher rank if it is either a strong seed itself or is one hop away from a strong seed via a high-affinity \textit{similar\_to} relation.

\textit{Ranking and selection.}
All candidates \(c\in\mathcal{C}\) are sorted in descending order of \( \rho(c) \). The score \( \rho(\cdot) \) is used solely to order cases for selection and does not modify the experience graph \( \mathcal{G} \).

\textit{Experience Hint Construction.}
Given the reranked case set \( \mathcal{C}_{\mathrm{rank}} \), EXG constructs a structured set of experience hints that summarizes relevant prior experience for the current attempt. Hint construction organizes cases by their semantic roles and selectively extracts salient information from each case.

\textit{Hint types from ranked cases.}
Given the reranked set \( \mathcal{C}_{\mathrm{rank}} \), EXG constructs three corresponding types of hints from the selected cases. Specifically, (i) \emph{fixed-by hints} are formed when a case participates in a \textit{fixed\_by} relation and thus provides explicit error--repair information; (ii) \emph{warning hints} are formed from unsuccessful cases to expose salient failure patterns; and (iii) \emph{golden hints} are formed from successful cases to provide concise positive exemplars. The hint type determines both the information extracted from each case and its priority during hint assembly.

\textit{Hint assembly.}
Hints are assembled by iterating over \( \mathcal{C}_{\mathrm{rank}} \) in descending order of relevance and instantiating the corresponding hint type for each case, subject to a limited budget. 
Fixed-by hints are added with the highest priority, followed by warning hints and golden hints. The resulting ordered hint list is denoted by
\begin{equation}
\mathcal{H} = \{ h_1, h_2, \dots, h_L \}.
\end{equation}

\textit{Hint usage.}
The constructed hint set \( \mathcal{H} \) is provided as auxiliary context for the current attempt as shown in Figure~\ref{fig:prompt_hint}.

\begin{figure}[t]
\centering
\begin{tikzpicture}[
    node distance=0cm,
    % 通用盒子样式：舒展平衡版
    box/.style={
        rectangle, 
        draw=black!70, 
        thick,
        align=left,
        text width=0.95\columnwidth, 
        % 1. 行距恢复自然：0.96倍
        font=\sffamily\scriptsize\linespread{0.96}\selectfont, 
        % 2. 内边距调大：4.2pt (让文字有呼吸感)
        inner sep=4.2pt, 
        outer sep=0pt
    },
    % 标题样式
    header_style/.style={
        font=\sffamily\bfseries\scriptsize, 
        minimum height=0.38cm, % 高度微增
        text depth=0.5pt,
        inner ysep=2pt 
    },
    % 颜色定义
    sys_c/.style={fill=cyan!20},
    usr_c/.style={fill=yellow!25},
    exg_c/.style={fill=red!15}
]

    % --- 1. System Part ---
    \node (sys_h) [box, sys_c, header_style] {System:};
    \node (sys_b) [box, fill=cyan!6, below=0pt of sys_h] {
        You are a coding assistant. Output ONLY Python code for the completion part. Do not include explanations.
    };

    % --- 2. User Part ---
    \node (usr_h) [box, usr_c, header_style, below=0pt of sys_b] {User:};
    \node (usr_b) [box, fill=yellow!6, below=0pt of usr_h] {
        Complete the following Python function:\\[0.5ex] % 加回了微小间距
        \texttt{<problem\_prompt>}
    };

    % --- 3. EXG Memory Hints ---
    \node (hint_h) [box, exg_c, header_style, below=0pt of usr_b] {=== MEMORY HINTS (via EXG) ===};
    \node (hint_b) [box, fill=red!6, below=0pt of hint_h] {
        \textit{<structured hints from past cases>}
    };

    % --- 4. Instruction ---
    \node (inst) [box, fill=white, below=0pt of hint_b, font=\sffamily\itshape\scriptsize] {
        Return only the code that should be appended after the prompt.
    };

\end{tikzpicture}
\vspace{-2mm}
\caption{EXG structured prompt architecture.}
\vspace{-3mm}
\label{fig:prompt_hint}
\end{figure}
\section{Self-Evolution with Experience Graphs}
\subsection{Online Self-Evolution}
Online self-evolution in EXG is realized through a closed-loop interaction between agent execution and the experience graph, as illustrated in Figure~\ref{fig1} (c). 
For each incoming task \( \tau \), the agent first initializes a provisional case \( c_q \), which contains the task input and contextual information but does not yet include an output or correctness outcome. 
Conditioned on \( c_q \), the experience graph \( \mathcal{G} \) is queried to retrieve relevant prior cases using the graph retrieval and reranking procedures defined in the EXG design.

Based on the reranked cases, EXG constructs a set of structured experience hints, which are injected into the agent’s prompt to guide the current attempt. 
The agent then executes the task under the guidance of these hints and produces an output. 
After execution completes, the provisional case \( c_q \) is finalized into a complete case by attaching the generated output, the correctness signal, and extracted execution signatures, yielding a new case \( c_{\tau}^{(k)} \).

The finalized case is then incorporated back into the experience graph according to the graph update rules. 
Specifically, a new case node corresponding to \( c_{\tau}^{(k)} \) is added to \( \mathcal{G} \) and connected to the task anchor node \( a_\tau \) via a \textit{contain} edge. 
If the new case corresponds to a corrective attempt following an earlier failure on the same task, a directed \textit{fixed\_by} edge is added to explicitly encode the error--repair relation. 
In addition, similarity relations between the new case and existing cases are established through \textit{similarity} edges based on their semantic affinity.

Through this self-evolving loop, experience generated during online interaction is incrementally externalized into the experience graph and becomes immediately available for subsequent tasks. As a result, online self-evolution in EXG enables experience to compound across tasks during deployment, while operating entirely at inference time without modifying the model parameters of agents.

\subsection{Offline Reuse}

\begin{figure}[t]
    \centering
    \includegraphics[width=\columnwidth]{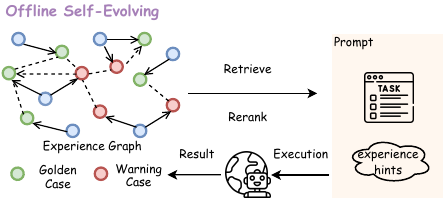}
    \caption{Offline self-evolving via graph reuse. EXG is pre-constructed during online process and held fixed.}
    \vspace{-2mm}
    \label{fig:offline}
\end{figure}

In addition to online self-evolution, EXG naturally supports an offline setting through reuse of a previously constructed experience graph. In this regime, the experience graph is built in advance from a set of training tasks and then held fixed during evaluation on unseen tasks, as illustrated in Figure~\ref{fig:offline}.

The offline setting differs from the online setting only in how the graph is deployed, rather than in how experience is represented or exploited. The same case abstraction, graph structure, and inference-time pipeline—including experience retrieval, relevance propagation, reranking, and hint construction—are used without modification, ensuring that experience is accessed and injected in a consistent manner across settings.

The key distinction is that no graph updates are performed in the offline setting. During evaluation, newly generated cases are not added to the graph, and no new similarity or corrective edges are created. As a result, the agent benefits from accumulated experience encoded in the frozen graph while preventing information leakage from test tasks.

By sharing a unified abstraction and inference-time pipeline across online and offline regimes, EXG provides a single experience-centric framework that supports both continual accumulation and static reuse, differing only in deployment rather than in mechanism.

\subsection{Plug-and-Play Integration}
Beyond serving as a standalone framework, EXG is designed as a plug-and-play experience module that can be integrated into a wide range of existing self-evolving agents. This property arises from the fact that EXG operates entirely at inference time and externalizes experience into a structured graph, without requiring changes to the underlying language model or task-specific solving procedures. As a result, EXG can be attached to different agent architectures as a shared experience layer, rather than as a tightly coupled algorithmic component.

In practice, EXG can augment both online and offline self-evolving methods by providing a persistent and structured experience foundation. For online agents, experience signals such as reflections or corrective feedback can be accumulated and reused across tasks, reducing repeated LLM calls for rediscovery. For offline agents, experience collected during training can be organized into the same graph structure and later accessed during evaluation through graph-based retrieval. In both cases, EXG enables experience to generalize across tasks via semantic and relational organization, offering a unified and reusable experience backbone that improves effectiveness while lowering inference-time cost.
\section{Experiments}
\subsection{Experimental Setup}
\subsubsection{Datasets and Models}
We evaluate EXG on a diverse set of benchmarks that cover program synthesis and multi-hop question answering, representing different forms of sequential reasoning and decision-making. Concretely, our evaluation includes HumanEval \cite{chen2021evaluatinglargelanguagemodels}, EvalPlus \cite{NEURIPS2023_43e9d647}, MuSiQue \cite{trivedi2022musique} and HotpotQA \cite{yang-etal-2018-hotpotqa}, each reflecting distinct problem settings and reasoning demands. We conduct experiments using multiple sizes of the Qwen3 \cite{yang2025qwen3technicalreport} model family. More details can be found in appendix~\ref{data}.

\subsubsection{Experimental Protocol and Hyperparameters}
Each task is allowed up to two attempts, consisting of a single initial attempt followed by at most one retry. Given a task query, EXG retrieves candidate seeds using dense vector similarity via a FAISS index. 
Specifically, the top-\(K_s\) most similar cases (\(K_s = 10\)) are first retrieved using sentence-level embeddings computed with a MiniLM encoder~\cite{NEURIPS2020_3f5ee243}. 
These seeds are then expanded through the experience graph by one-hop traversal over \textit{similarity} edges with a fanout \(F_{\text{sim}} = 5\). Each task is associated with an anchor node, from which at most one directly related case is selected, followed by similarity-based bridge expansion with a fanout \(F_{\text{bridge}} = 5\). After merging and deduplication, the total number of candidate cases is globally capped at \(K_c = 30\). Candidate relevance is computed by combining prompt-level similarity and failure-aware similarity using a weighting factor \(\alpha = 0.8\). From the ranked candidates, EXG constructs at most \(H = 5\) experience hints, which are injected into the model prompt. Effectiveness is evaluated using pass@1 and pass@2, while efficiency is measured by the avarage number of LLM calls, LLM inference latency, and retrieval latency.

\subsection{Online Experience Graph Performance}

\newcommand{\mhead}[1]{\makebox[1.6cm][c]{\textbf{#1}}}

\begin{table*}[t]
\centering
\caption{Online performance comparison across baselines and EXG, reported in terms of pass@1 and pass@2 on different datasets and models.}
\label{tab:online_perf}
\setlength{\tabcolsep}{4.0pt}
\renewcommand{\arraystretch}{1.15}
\small

\begin{tabular}{
  c c
  S[table-format=1.2] S[table-format=1.2]
  S[table-format=1.2] S[table-format=1.2]
  S[table-format=1.2] S[table-format=1.2]
  S[table-format=1.2] S[table-format=1.2]
  S[table-format=1.2] S[table-format=1.2]
  S[table-format=1.2] S[table-format=1.2]
  S[table-format=1.2] S[table-format=1.2]
}
\toprule
\textbf{Dataset} & \textbf{Model}
& \multicolumn{2}{c}{\mhead{Reflexion}}
& \multicolumn{2}{c}{\mhead{SE-Agent}}
& \multicolumn{2}{c}{\mhead{SE-Agent-Rev}}
& \multicolumn{2}{c}{\mhead{EXG}}
& \multicolumn{2}{c}{\mhead{EXG-Reflexion}}
& \multicolumn{2}{c}{\mhead{EXG-SE}}
& \multicolumn{2}{c}{\mhead{EXG-SE-Rev}} \\
\cmidrule(lr){3-4}\cmidrule(lr){5-6}\cmidrule(lr){7-8}\cmidrule(lr){9-10}
\cmidrule(lr){11-12}\cmidrule(lr){13-14}\cmidrule(lr){15-16}
&
& \multicolumn{1}{c}{p@1} & \multicolumn{1}{c}{p@2}
& \multicolumn{1}{c}{p@1} & \multicolumn{1}{c}{p@2}
& \multicolumn{1}{c}{p@1} & \multicolumn{1}{c}{p@2}
& \multicolumn{1}{c}{p@1} & \multicolumn{1}{c}{p@2}
& \multicolumn{1}{c}{p@1} & \multicolumn{1}{c}{p@2}
& \multicolumn{1}{c}{p@1} & \multicolumn{1}{c}{p@2}
& \multicolumn{1}{c}{p@1} & \multicolumn{1}{c}{p@2} \\
\midrule

\multirow{3}{*}{\centering HumanEval}
& \centering Qwen3-1.7B
& 0.207 & 0.543  & 0.195 & 0.524  & 0.207 & 0.561
& 0.537 & 0.585  & \textbf{0.573} & \textbf{0.695}  & 0.512 & 0.610  & 0.537 & 0.598 \\
& \centering Qwen3-8B
& 0.201 & 0.610  & 0.177 & 0.720  & 0.189 & 0.823
& 0.415 & 0.488  & 0.506 & 0.720  & 0.573 & \textbf{0.835}  & \textbf{0.591} & 0.799 \\
& \centering Qwen3-Coder-Flash
& 0.585 & 0.780  & 0.598 & 0.939  & 0.640 & 0.945
& 0.805 & 0.835  & 0.817 & 0.872  & \textbf{0.860} & \textbf{0.951}  & 0.835 & 0.921 \\
\midrule

\multirow{3}{*}{\centering EvalPlus}
& \centering Qwen3-1.7B
& 0.134 & 0.342  & 0.189 & 0.207  & 0.207 & 0.482
& 0.323 & 0.439  & 0.317 & \textbf{0.494}  & \textbf{0.408} & 0.415  & 0.299 & 0.445 \\
& \centering Qwen3-8B
& 0.183 & 0.274  & 0.317 & 0.341  & 0.341 & \textbf{0.750}
& 0.409 & 0.530  & 0.390 & 0.598  & \textbf{0.524} & 0.537  & 0.311 & 0.677 \\

& \centering Qwen3-Coder-Flash
& 0.585 & 0.872  & 0.610 & 0.768  & 0.591 & 0.848
& 0.811 & 0.866  & 0.793 & 0.866  & 0.793 & 0.848  & \textbf{0.817} & \textbf{0.915} \\
\midrule

\multirow{3}{*}{\centering MuSiQue}
& \centering Qwen3-14B
& 0.174 & 0.326  & 0.356 & 0.700  & 0.382 & 0.836
& 0.320 & 0.406 & 0.480 & 0.748 & 0.490 & 0.736    & \textbf{0.560} & \textbf{0.896} \\
& \centering Qwen-Plus
& 0.542 & 0.940  & 0.552 & 0.840  & 0.568 & 0.970
& 0.608 & 0.672  & 0.652 & 0.946  & 0.652 & 0.854  & \textbf{0.694} & \textbf{0.990} \\
& \centering Qwen-Max
& 0.346 & 0.692  & 0.348 & 0.746  & 0.370 & 0.882
& 0.446 & 0.516  & 0.478 & 0.762  & 0.486 & \textbf{0.926}  & \textbf{0.518} & 0.924 \\
\midrule

\multirow{3}{*}{\centering HotpotQA}
& \centering Qwen3-8B
& 0.283 & 0.494  & 0.283 & 0.410  & 0.275 & 0.344
& 0.460 & 0.500  & \textbf{0.462} & \textbf{0.557}  & 0.445 & 0.516  & 0.458 & 0.509 \\
& \centering Qwen-Plus
& 0.597 & 0.662  & 0.603 & 0.670  & 0.587 & 0.669
& \textbf{0.618} & 0.642  & 0.601 & 0.674  & 0.613 & \textbf{0.681}  & 0.608 & 0.680 \\
& \centering Qwen-Max
& 0.616 & 0.686  & 0.607 & 0.669  & 0.621 & 0.680 & \textbf{0.659} & 0.684  & 0.651 & 0.686  & 0.657 & \textbf{0.707}  & 0.655 & 0.703 \\
\bottomrule
\end{tabular}
\end{table*}

\subsubsection{Baselines}
We compare EXG against representative online self-evolving agents, including Reflexion and SE-Agent. Reflexion \cite{NEURIPS2023_1b44b878} augments the agent with self-generated verbal feedback, enabling iterative correction by reflecting on previous failures, while SE-Agent \cite{lin2025seagentselfevolutiontrajectoryoptimization} performs trajectory-level self-evolution through operations such as revision and recombination across multiple reasoning paths. Since the benchmarks considered in this work typically involve short interaction horizons and limited trajectories, we additionally include a simplified variant of SE-Agent that only applies the revision operation denoted as SE-Agent-Rev, in order to provide a more comparable baseline under constrained retry budgets.

To ensure a fair comparison under identical model interaction constraints, we further integrate EXG into these online self-evolving frameworks in a plug-and-play manner, resulting in EXG-Reflexion and EXG-SE variants. These settings allow us to evaluate whether explicitly structuring experience as a graph can complement existing online self-evolution mechanisms without increasing the number of model calls or modifying the underlying agent logic.

\subsubsection{Results}
Table \ref{tab:online_perf} shows that EXG consistently strengthens online self-evolution across both code generation and multi-hop reasoning tasks. Compared with Reflexion and SE-Agent variants, EXG together with its plug-and-play extensions achieve higher pass@1 in nearly all settings, indicating more effective first-attempt reasoning. These improvements generally translate to higher pass@2 as well, with the best pass@2 results often achieved by EXG-based methods, suggesting that accumulated experience not only reduces repeated errors but also increases the utility of the limited retry budget. Importantly, these gains hold across model scales, from compact models to stronger code- and reasoning backbones.

On HumanEval and EvalPlus, EXG delivers the largest relative gains on smaller and mid-sized models, where prior experience can most directly compensate for limited model capacity. For example, on HumanEval with Qwen3-1.7B, EXG improves pass@1 by more than \textbf{150\%} relative to Reflexion, while EXG-Reflexion further increases pass@2 by roughly \textbf{30\%}. As model capacity increases, the relative gains narrow but remain consistent: with Qwen3-Coder-Flash, EXG still improves pass@1 by over \textbf{30\%} compared to the strongest non-graph baseline. On EvalPlus, which applies stricter correctness checks, EXG maintains comparable or larger relative improvements in pass@1 while also improving pass@2, indicating that the benefits are not due to overfitting shallow test cases but reflect more robust semantic correctness.

On MuSiQue and HotpotQA, EXG exhibits a different but complementary pattern. On MuSiQue, which emphasizes compositional reasoning, EXG substantially boosts pass@1, with relative improvements of around \textbf{40–50\%} on Qwen3-14B and close to \textbf{30\%} on Qwen-Plus, while also yielding consistent relative gains in pass@2 across model variants. This suggests that graph-structured experience helps abstract reusable reasoning patterns across related question compositions. On HotpotQA, where question diversity and reasoning paths are less regular, gains are smaller but stable: EXG improves pass@1 by over \textbf{60\%} on Qwen3-8B and by around \textbf{7\%} on Qwen-Max. Taken together, these results indicate that EXG is particularly effective when tasks share latent structural regularities, while still providing consistent first-attempt improvements in more heterogeneous reasoning settings.

\subsection{Online Efficiency Analysis}
\subsubsection{LLM Calls}

\begin{figure}[t]
  \centering
  \includegraphics[width=0.9\columnwidth]{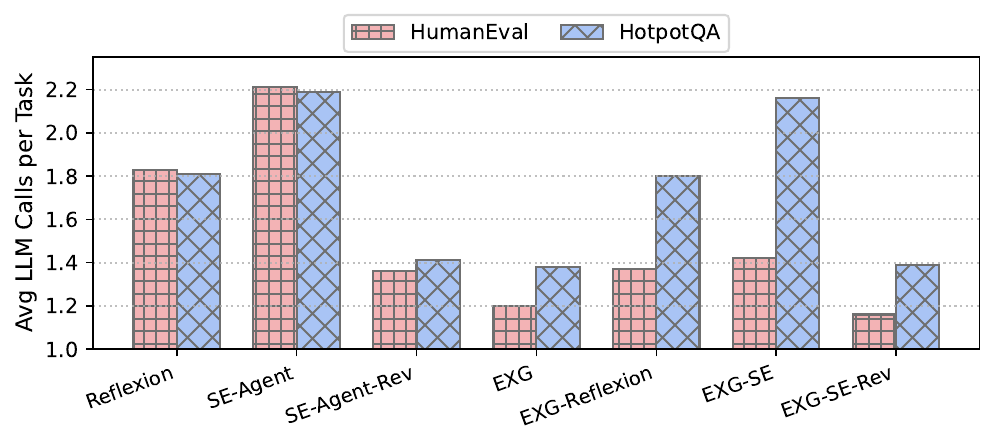}
  \caption{Average number of LLM calls per task under the online setting on HumanEval and HotpotQA.}
  \label{fig:llm_calls}
\end{figure}

Figure~\ref{fig:llm_calls} illustrates the average number of LLM calls per task for different online methods. On HumanEval, the base EXG method requires 1.20 calls per task, compared to 1.83 for Reflexion and 2.21 for SE-Agent, corresponding to reductions of \textbf{34.4\%} and \textbf{45.7\%}, respectively. EXG also consistently outperforms revised baselines, reducing LLM calls from 1.36 to 1.20, a relative decrease of \textbf{11.8\%}. On HotpotQA, the reduction remains consistent with EXG lowering LLM calls from 2.19 for the most expensive baseline to 1.38, achieving a \textbf{37\%} reduction.

Overall, EXG-based methods require fewer LLM calls than their corresponding online baselines, indicating that the observed performance improvements are achieved with lower interaction cost rather than increased model usage. This reduction primarily stems from improved first-attempt success rates: unlike Reflexion or SE-Agent, which invoke the LLM multiple times for reflection or revision after failures, EXG provides richer guidance at inference time through structured experience reuse, thereby avoiding repeated model calls.

This trend also holds for plug-and-play variants. On HumanEval, EXG-Reflexion reduces the average number of LLM calls from \textbf{1.83} to \textbf{1.37}, while EXG-SE-Rev achieves the lowest cost at \textbf{1.16} calls per task. On HotpotQA, plug-and-play variants likewise incur fewer LLM calls than their original counterparts, although the reductions are more moderate due to the greater diversity of question types and reasoning paths.

\subsubsection{Latency}

\begin{figure}[t]
  \centering
  \includegraphics[width=0.9\columnwidth]{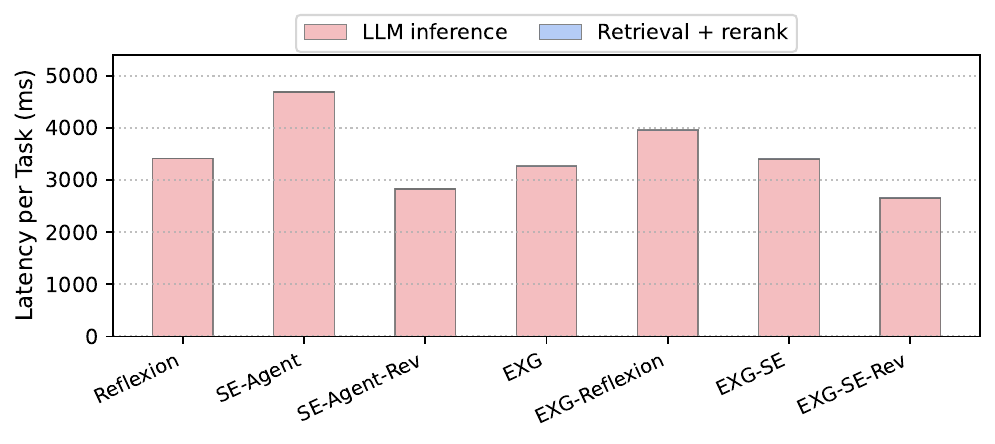}
  \caption{Latency breakdown under the online setting on HumanEval, showing LLM inference and retrieval overhead.}
  \label{fig:latency_breakdown}
\end{figure}

Figure~\ref{fig:latency_breakdown} reports the latency breakdown of different online methods on HumanEval, decomposing per-task runtime into LLM inference time and auxiliary overhead. In general, EXG-based methods achieve lower or comparable end-to-end latency, while the retrieval overhead introduced by the experience graph is minimal (\textbf{\(\sim\)18--22\,ms}). Concretely, EXG reduces the average LLM latency to $3,259$\,ms, compared to $3,416$\,ms for Reflexion and $4,689$\,ms for SE-Agent, corresponding to latency reductions of \textbf{4.6\%} and \textbf{30.5\%}, respectively, by avoiding repeated reflection calls. 
When combined with reflection, EXG-Reflexion incurs slightly higher latency than Reflexion alone ($3,943$\,ms vs.\ $3,416$\,ms) due to the additional reflection step. 
For SE-Agent, EXG consistently reduces latency across variants: EXG-SE lowers LLM inference time from $4,689$\,ms to $3,399$\,ms (\textbf{27.5\%} reduction), while EXG-SE-Rev further reduces latency relative to SE-Agent-Rev from $2,826.6$\,ms to $2,648$\,ms, corresponding to a \textbf{6.3\%} reduction. These results indicate that EXG effectively reduces reflection-heavy computation while introducing only negligible retrieval overhead.

\newcommand{\bhead}[1]{\makebox[1.5cm][c]{\textbf{#1}}}
\newcommand{\colhead}[1]{\makebox[0.8cm][c]{#1}}
\newcommand{\callscol}[1]{\makebox[1cm][c]{#1}}
\newcommand{\bench}[2]{%
  \begin{tabular}[c]{@{}c@{}}
    \textbf{#1}\\[-3pt]
    \textbf{(#2)}
  \end{tabular}
}

\begin{table*}[t]
\centering
\caption{Offline performance comparison between ExpeL and EXG-Reflexion. Results include collection-stage pass@1 and pass@2 (C-p@1 and C-p@2, resp.) with average LLM calls, and test-stage pass@1 and pass@2 (T-p@1 and T-p@2, resp.).}
\label{tab:offline_reuse_wide}
\setlength{\tabcolsep}{4pt}
\renewcommand{\arraystretch}{1}
\small

% 1 benchmark column + 10 equal-width metric columns
\begin{tabular}{
  c
  *{10}{>{\centering\arraybackslash}p{0.9cm}}
}
\toprule
\textbf{Benchmark}
& \multicolumn{5}{c}{\bhead{ExpeL}}
& \multicolumn{5}{c}{\bhead{EXG-Reflexion}} \\
\cmidrule(lr){2-6}\cmidrule(lr){7-11}
& \colhead{C-p@1} & \colhead{C-p@2} & \callscol{Avg. Calls} & \colhead{T-p@1} & \colhead{T-p@2}
& \colhead{C-p@1} & \colhead{C-p@2} & \callscol{Avg. Calls} & \colhead{T-p@1} & \colhead{T-p@2} \\
\midrule

\bench{HumanEval}{Qwen3-Coder-Flash}
& 0.573 & 0.771 & 1.85 & 0.909 & 0.909
& 0.824 & 0.885 & 1.35 & 0.879 & 0.879 \\

% \addlinespace[1pt]

\bench{HotpotQA}{Qwen-Plus}
& 0.600 & 0.661 & 1.80 & 0.590 & 0.633
& 0.614 & 0.670 & 1.79 & 0.610 & 0.637 \\

\bottomrule
\end{tabular}
\end{table*}

\subsection{Offline Experience Graph Performance}
\subsubsection{Baselines}
We adopt ExpeL \cite{10.1609/aaai.v38i17.29936} as the representative offline self-evolving baseline and compare it with EXG-Reflexion under a matched two-stage protocol for fair comparison.
ExpeL performs experience collection in an online phase using reflexion-based retries, and in the offline phase it processes the collected data, particularly the reflection contents, to abstract a set of high-level insights. 
To align with this mechanism, we choose the reflexion variant of EXG for offline comparison for reasons below.
(1) EXG-Reflexion also relies on reflection-based signals during online collection.
(2) In the offline stage, the experience graph constructed is kept fixed, and insights are extracted from the graph together with the associated reflections, matching ExpeL’s offline abstraction process. 
At test time, each method follows its own retrieval strategy: ExpeL injects a fixed memory consisting of five insights and five golden cases, whereas EXG-Reflexion injects five insights together with five experience hints retrieved from the graph, which may include golden, warning, and fix-related cases. For each benchmark, the dataset is randomly shuffled and split into a 7:3 ratio for online collection and offline evaluation, respectively.

\subsubsection{Results}
As shown in Table~\ref{tab:offline_reuse_wide}, during the online collection stage, EXG-Reflexion consistently achieves stronger performance with lower interaction cost than ExpeL. On HumanEval, EXG-Reflexion improves collection-stage pass@1 from \textbf{0.573} to \textbf{0.824} and pass@2 from \textbf{0.771} to \textbf{0.885}, while reducing the average number of LLM calls from \textbf{1.85} to \textbf{1.35}. This substantial gain indicates that the experience graph enables EXG-Reflexion to solve a larger fraction of tasks on the first attempt, significantly reducing reliance on repeated reflections during online collection. In the offline test stage with a frozen memory, the two methods exhibit comparable performance on HumanEval. 
EXG-Reflexion attains a test-stage pass@1 of \textbf{0.879}, slightly below ExpeL’s \textbf{0.909}.
This gap can be partially attributed to the reduced volume of reflection traces collected online by EXG-Reflexion: because EXG more effectively leverages experience during online interaction, fewer reflection calls are triggered, resulting in fewer reflection-derived insights available for offline reuse. On HotpotQA, the task distribution is less concentrated and experience reuse is inherently more challenging. During online collection, EXG-Reflexion achieves slightly higher pass@1 (\textbf{0.614} vs.\ \textbf{0.600}) and pass@2 (\textbf{0.670} vs.\ \textbf{0.661}) than ExpeL, while incurring a comparable number of LLM calls (\textbf{1.79} vs.\ \textbf{1.80}). In the offline test stage, EXG-Reflexion continues to outperform ExpeL, achieving pass@1 of \textbf{0.610} and pass@2 of \textbf{0.637}, compared to \textbf{0.590} and \textbf{0.633}, respectively. In summary, these results indicate that EXG can effectively construct a reusable and generalizable experience graph during online interaction, which supports competitive and, in some cases, improved performance in offline evaluation despite operating under a frozen-memory setting.

\subsection{Ablation Study}
\subsubsection{Structural Ablations of Experience Graph}
\begin{figure}[t]
    \centering
    \includegraphics[width=0.8\columnwidth]{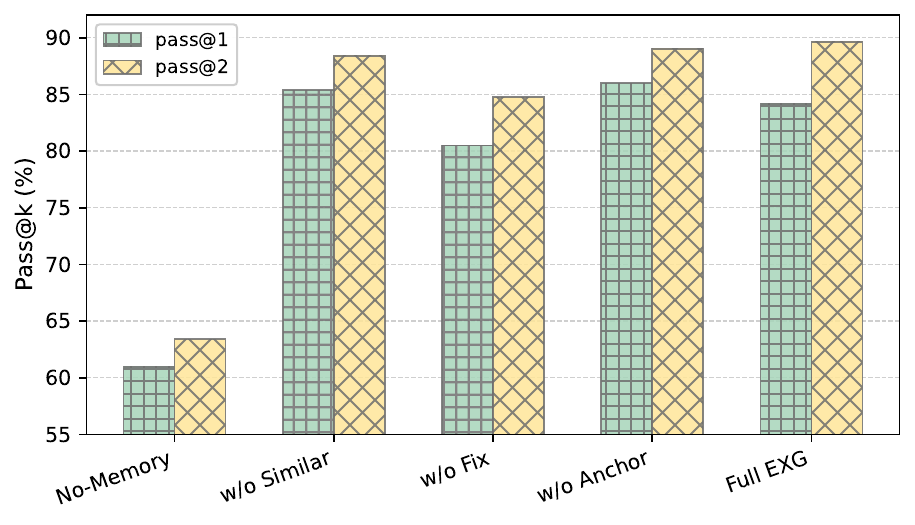}
    \caption{Structural ablation study of EXG on HumanEval under different graph configurations.}
    \label{fig:structural_ablation}
\end{figure}
We conduct a structural ablation study to examine the contribution of different components in the experience graph. Specifically, we compare the full EXG against several variants that remove individual structural elements. 

\textbf{No-Memory} disables experience graph construction entirely.

\textbf{w/o Similar} removes the \textit{similarity} edges from the graph and replaces them with standard query-based similarity retrieval.

\textbf{w/o Fix} removes the \textit{correction} edges, eliminating explicit error--repair links between failed and successful cases.

\textbf{w/o Anchor} removes task-level anchor nodes, forcing retrieval to rely solely on case-level relations without task-centric grouping.

\textbf{Full EXG} retains all node and edge types, including task anchors, \textit{similarity} edges, and \textit{correction} edges.

As shown in Figure~\ref{fig:structural_ablation}, the full EXG consistently achieves the best performance, while No-Memory performs worst due to the absence of structured experience. Among the ablated variants, removing \textit{correction} edges leads to the largest performance drop, indicating that explicit corrective information from failures to successful solutions provides a stable and significant benefit. 
Removing \textit{similarity} edges also degrades performance, suggesting that similarity relations play an important role in retrieving relevant cases. Finally, removing anchor nodes results in a smaller but consistent decrease compared to the full EXG, showing that task-level anchoring and bridge-based expansion further improve retrieval effectiveness. Together, these results demonstrate that the performance gains of EXG arise from the joint contribution of multiple graph components, with \textit{fixed\_by} relations providing the most critical signal.

\subsubsection{Hint Type Ablation}
We further conduct a hint-type ablation study to analyze the impact of different compositions of experience hints. As shown in Table~\ref{tab:hint_ablation}, No-Memory serves as a reference without experience injection, while Similar-Only injects only golden cases retrieved by similarity, providing correct but limited guidance. Warning-Only further enriches the injected information by incorporating warning cases that encode failure types and signatures, enabling the agent to reason about potential mistakes. The full EXG setting injects a heterogeneous set of experience hints, including corrective signals derived from \textit{fixed\_by} relations in addition to golden and warning information. The results show a clear performance improvement as the injected experience becomes more diverse, with EXG achieving the best pass@1 and pass@2. This indicates that, under the EXG framework, correct solutions, failure-aware signals, and explicit corrective information each provide complementary benefits, and their joint composition is crucial for effective experience reuse.

\begin{table}[t]
\centering
\caption{Hint type ablation study of EXG on HumanEval.}
\label{tab:hint_ablation}
\setlength{\tabcolsep}{7pt}
\renewcommand{\arraystretch}{1}
\small
\begin{tabular}{lcc}
\toprule
\textbf{Group} & \textbf{pass@1} & \textbf{pass@2} \\
\midrule
No-Memory     & 0.610 & 0.640 \\
Similar-Only  & 0.823 & 0.835 \\
Warning-Only  & 0.866 & 0.884 \\
\midrule
\textbf{EXG} & \textbf{0.872} & \textbf{0.896} \\
\bottomrule
\end{tabular}
\vspace{-3mm}
\end{table}

\section{Related Work}
\subsection{Self-Evolving Agents}
Self-evolving agents seek to improve their behavior over time by leveraging experience generated through their own interactions, rather than remaining functionally fixed. Prior work in this area can be broadly categorized by whether self-evolution is achieved through updating model parameters or without modifying them. Parameter-updating approaches \cite{wu2025evolverselfevolvingllmagents,zhang2025agentlearningearlyexperience,xia2025agent0unleashingselfevolvingagents,zhang2026darwintodllmdrivenlifelong,yue2026drzeroselfevolvingsearch,lu2025searchselfplaypushingfrontier} internalize experience via iterative training or reinforcement learning, enabling continual improvement at the cost of retraining overhead and reduced modularity. In contrast, non-parameter methods realize self-evolution at inference time and can be further divided based on how experience is utilized. Online approaches \cite{lin2025seagentselfevolutiontrajectoryoptimization,NEURIPS2023_1b44b878,zhai2025agentevolverefficientselfevolvingagent} primarily rely on reflexion or trajectory-level optimization to correct behavior within a single task, resulting in task-local improvements without persistent experience reuse. Offline approaches \cite{cao2025remembermerefineme, 10.1609/aaai.v38i17.29936,liu-etal-2025-contextual,zhang2025agenticcontextengineeringevolving,cai2025flexcontinuousagentevolution,yang2025learningjobexperiencedrivenselfevolving,10.1016/j.neucom.2025.130470,hassell2025learningsupervisionsemanticepisodic,zhang2026memrlselfevolvingagentsruntime,liu2025trainingenablingselfevolutionagents} instead accumulate experience across tasks and apply consolidation or abstraction before reuse, enabling cross-task improvement through external memory or experience repositories. 
Memory-based methods highlight the role of persistent experience in self-evolution. However, they typically use flat or weakly structured representations and require substantial offline refinement before reuse.

\subsection{Agent Memory Systems}
Agent memory systems investigate how external memory can be incorporated to support long-horizon interaction while improving consistency and efficiency in LLM-based agents. A subset of this line of work \cite{zheng2024synapsetrajectoryasexemplarpromptingmemory,wang2024agentworkflowmemory} is closely related to memory-based approaches in self-evolving agents, since they also reuse past trajectories, experiences, or reasoning traces across tasks; however, such methods primarily treat memory as an auxiliary information source for task execution, focusing on retrieval and prompt injection rather than modeling how an agent’s behavior systematically evolves with accumulated experience. Beyond this overlap, agent memory systems constitute a broader and more mature research direction, encompassing methods that model persistent context or knowledge for long-term coherence \cite{yu2026agenticmemorylearningunified,xu2025amemagenticmemoryllm,westhäußer2025enablingpersonalizedlongterminteractions,fang2025lightmemlightweightefficientmemoryaugmented,wei2025mlpmemoryretrieverpretrainedmemory}, treat memory as a system-level or control component governing memory organization and lifecycle \cite{hu2026evermemosselforganizingmemoryoperating,qian2026memobrainexecutivememoryagentic,zhang2025gmemorytracinghierarchicalmemory,zhang2025memevolvemetaevolutionagentmemory}, and conduct efficiency analysis and benchmarking to examine the computational trade-offs and behavioral effects of memory in continual settings \cite{ai2025memorybenchbenchmarkmemorycontinual,wei2025evomemorybenchmarkingllmagent,xiong2025memorymanagementimpactsllm,10.1145/3748302}. Accordingly, approaches that aim to explicitly model experience-driven agent evolution do not fall within the scope of memory systems, though techniques developed in this literature provide valuable insights that can be leveraged to support more effective self-evolving agents.
\section{Conclusion}
In this work, we introduced EXG, an experience graph framework that enables self-evolving agents to systematically accumulate, organize, and reuse experience during deployment without modifying model parameters. By externalizing experience into a structured graph and operating entirely at inference time, EXG provides a principled mechanism for transforming ephemeral trial-and-error interactions into reusable knowledge. Extensive experiments across online and offline settings demonstrate that EXG improves both effectiveness and efficiency: it achieves higher task success rates with fewer model calls during online interaction and retains competitive performance when reused as a frozen experience module offline. Moreover, EXG reduces overall LLM inference time by mitigating unnecessary retries and reflections, while the overhead introduced by graph-based retrieval remains marginal. Through comprehensive ablation studies, we further show that EXG’s benefits arise from both its graph structure and its ability to compose heterogeneous experience signals in a type-aware manner. In addition, EXG can be readily incorporated into existing self-evolving agents as an external experience layer, consistently enhancing performance without requiring changes to the underlying agent architecture. Together, these results suggest that structuring experience as a graph offers a scalable and modular foundation for self-evolving agents, enabling improvements to compound over time rather than reset across tasks.

\bibliographystyle{ACM-Reference-Format}
\bibliography{ref}

@String{Computing = "Computing" }

@String{Computer = "{IEEE} Computer" }

@inproceedings{10.5555/3600270.3602070,
author = {Wei, Jason and Wang, Xuezhi and Schuurmans, Dale and Bosma, Maarten and Ichter, Brian and Xia, Fei and Chi, Ed H. and Le, Quoc V. and Zhou, Denny},
title = {Chain-of-thought prompting elicits reasoning in large language models},
year = {2022},
isbn = {9781713871088},
publisher = {Curran Associates Inc.},
address = {Red Hook, NY, USA},
booktitle = {Proceedings of the 36th International Conference on Neural Information Processing Systems},
articleno = {1800},
numpages = {14},
location = {New Orleans, LA, USA},
series = {NIPS '22}
}

@misc{yao2023reactsynergizingreasoningacting,
      title={ReAct: Synergizing Reasoning and Acting in Language Models}, 
      author={Shunyu Yao and Jeffrey Zhao and Dian Yu and Nan Du and Izhak Shafran and Karthik Narasimhan and Yuan Cao},
      year={2023},
      eprint={2210.03629},
      archivePrefix={arXiv},
      primaryClass={cs.CL},
      url={https://arxiv.org/abs/2210.03629}, 
}

@misc{wang2023selfconsistencyimproveschainthought,
      title={Self-Consistency Improves Chain of Thought Reasoning in Language Models}, 
      author={Xuezhi Wang and Jason Wei and Dale Schuurmans and Quoc Le and Ed Chi and Sharan Narang and Aakanksha Chowdhery and Denny Zhou},
      year={2023},
      eprint={2203.11171},
      archivePrefix={arXiv},
      primaryClass={cs.CL},
      url={https://arxiv.org/abs/2203.11171}, 
}

@misc{chen2021evaluatinglargelanguagemodels,
      title={Evaluating Large Language Models Trained on Code}, 
      author={Mark Chen and Jerry Tworek and Heewoo Jun and Qiming Yuan and Henrique Ponde de Oliveira Pinto and Jared Kaplan and Harri Edwards and Yuri Burda and Nicholas Joseph and Greg Brockman and Alex Ray and Raul Puri and Gretchen Krueger and Michael Petrov and Heidy Khlaaf and Girish Sastry and Pamela Mishkin and Brooke Chan and Scott Gray and Nick Ryder and Mikhail Pavlov and Alethea Power and Lukasz Kaiser and Mohammad Bavarian and Clemens Winter and Philippe Tillet and Felipe Petroski Such and Dave Cummings and Matthias Plappert and Fotios Chantzis and Elizabeth Barnes and Ariel Herbert-Voss and William Hebgen Guss and Alex Nichol and Alex Paino and Nikolas Tezak and Jie Tang and Igor Babuschkin and Suchir Balaji and Shantanu Jain and William Saunders and Christopher Hesse and Andrew N. Carr and Jan Leike and Josh Achiam and Vedant Misra and Evan Morikawa and Alec Radford and Matthew Knight and Miles Brundage and Mira Murati and Katie Mayer and Peter Welinder and Bob McGrew and Dario Amodei and Sam McCandlish and Ilya Sutskever and Wojciech Zaremba},
      year={2021},
      eprint={2107.03374},
      archivePrefix={arXiv},
      primaryClass={cs.LG},
      url={https://arxiv.org/abs/2107.03374}, 
}

@article{trivedi2022musique,
  title={MuSiQue: Multihop Questions via Single-hop Question Composition},
  author={Trivedi, Harsh and Balasubramanian, Niranjan and Khot, Tushar and Sabharwal, Ashish},
  journal={Transactions of the Association for Computational Linguistics},
  volume={10},
  pages={539--554},
  year={2022},
  url={https://aclanthology.org/2022.tacl-1.31/},
  doi={10.1162/tacl_a_00475}
}

@inproceedings{yang-etal-2018-hotpotqa,
    title = "{H}otpot{QA}: A Dataset for Diverse, Explainable Multi-hop Question Answering",
    author = "Yang, Zhilin  and
      Qi, Peng  and
      Zhang, Saizheng  and
      Bengio, Yoshua  and
      Cohen, William  and
      Salakhutdinov, Ruslan  and
      Manning, Christopher D.",
    editor = "Riloff, Ellen  and
      Chiang, David  and
      Hockenmaier, Julia  and
      Tsujii, Jun{'}ichi",
    booktitle = "Proceedings of the 2018 Conference on Empirical Methods in Natural Language Processing",
    month = oct # "-" # nov,
    year = "2018",
    address = "Brussels, Belgium",
    publisher = "Association for Computational Linguistics",
    url = "https://aclanthology.org/D18-1259/",
    doi = "10.18653/v1/D18-1259",
    pages = "2369--2380"
}

@inproceedings{10.1145/3586183.3606763,
author = {Park, Joon Sung and O'Brien, Joseph and Cai, Carrie Jun and Morris, Meredith Ringel and Liang, Percy and Bernstein, Michael S.},
title = {Generative Agents: Interactive Simulacra of Human Behavior},
year = {2023},
isbn = {9798400701320},
publisher = {Association for Computing Machinery},
address = {New York, NY, USA},
url = {https://doi.org/10.1145/3586183.3606763},
doi = {10.1145/3586183.3606763},
booktitle = {Proceedings of the 36th Annual ACM Symposium on User Interface Software and Technology},
articleno = {2},
numpages = {22},
location = {San Francisco, CA, USA},
series = {UIST '23}
}

@misc{packer2024memgptllmsoperatingsystems,
      title={MemGPT: Towards LLMs as Operating Systems}, 
      author={Charles Packer and Sarah Wooders and Kevin Lin and Vivian Fang and Shishir G. Patil and Ion Stoica and Joseph E. Gonzalez},
      year={2024},
      eprint={2310.08560},
      archivePrefix={arXiv},
      primaryClass={cs.AI},
      url={https://arxiv.org/abs/2310.08560}, 
}

@misc{wang2023voyageropenendedembodiedagent,
      title={Voyager: An Open-Ended Embodied Agent with Large Language Models}, 
      author={Guanzhi Wang and Yuqi Xie and Yunfan Jiang and Ajay Mandlekar and Chaowei Xiao and Yuke Zhu and Linxi Fan and Anima Anandkumar},
      year={2023},
      eprint={2305.16291},
      archivePrefix={arXiv},
      primaryClass={cs.AI},
      url={https://arxiv.org/abs/2305.16291}, 
}

@inproceedings{NEURIPS2023_91edff07,
 author = {Madaan, Aman and Tandon, Niket and Gupta, Prakhar and Hallinan, Skyler and Gao, Luyu and Wiegreffe, Sarah and Alon, Uri and Dziri, Nouha and Prabhumoye, Shrimai and Yang, Yiming and Gupta, Shashank and Majumder, Bodhisattwa Prasad and Hermann, Katherine and Welleck, Sean and Yazdanbakhsh, Amir and Clark, Peter},
 booktitle = {Advances in Neural Information Processing Systems},
 editor = {A. Oh and T. Naumann and A. Globerson and K. Saenko and M. Hardt and S. Levine},
 pages = {46534--46594},
 publisher = {Curran Associates, Inc.},
 title = {Self-Refine: Iterative Refinement with Self-Feedback},
 volume = {36},
 year = {2023},
 url = {https://proceedings.neurips.cc/paper_files/paper/2023/file/91edff07232fb1b55a505a9e9f6c0ff3-Paper-Conference.pdf}
}

@inproceedings{NEURIPS2023_1b44b878,
 author = {Shinn, Noah and Cassano, Federico and Gopinath, Ashwin and Narasimhan, Karthik and Yao, Shunyu},
 booktitle = {Advances in Neural Information Processing Systems},
 editor = {A. Oh and T. Naumann and A. Globerson and K. Saenko and M. Hardt and S. Levine},
 pages = {8634--8652},
 publisher = {Curran Associates, Inc.},
 title = {Reflexion: language agents with verbal reinforcement learning},
 volume = {36},
 year = {2023},
 url = {https://proceedings.neurips.cc/paper_files/paper/2023/file/1b44b878bb782e6954cd888628510e90-Paper-Conference.pdf}
}

@inproceedings{ICLR2024_3339f19c,
 author = {Yang, Chengrun and Wang, Xuezhi and Lu, Yifeng and Liu, Hanxiao and Le, Quoc V and Zhou, Denny and Chen, Xinyun},
 booktitle = {International Conference on Learning Representations},
 editor = {B. Kim and Y. Yue and S. Chaudhuri and K. Fragkiadaki and M. Khan and Y. Sun},
 pages = {12028--12068},
 title = {Large Language Models as Optimizers},
 volume = {2024},
 year = {2024},
 url = {https://proceedings.iclr.cc/paper_files/paper/2024/file/3339f19c5fcee3ad74502947a32be9e6-Paper-Conference.pdf}
}

@misc{wu2025evolverselfevolvingllmagents,
      title={EvolveR: Self-Evolving LLM Agents through an Experience-Driven Lifecycle}, 
      author={Rong Wu and Xiaoman Wang and Jianbiao Mei and Pinlong Cai and Daocheng Fu and Cheng Yang and Licheng Wen and Xuemeng Yang and Yufan Shen and Yuxin Wang and Botian Shi},
      year={2025},
      eprint={2510.16079},
      archivePrefix={arXiv},
      primaryClass={cs.CL},
      url={https://arxiv.org/abs/2510.16079}, 
}

@misc{lin2025seagentselfevolutiontrajectoryoptimization,
      title={SE-Agent: Self-Evolution Trajectory Optimization in Multi-Step Reasoning with LLM-Based Agents}, 
      author={Jiaye Lin and Yifu Guo and Yuzhen Han and Sen Hu and Ziyi Ni and Licheng Wang and Mingguang Chen and Hongzhang Liu and Ronghao Chen and Yangfan He and Daxin Jiang and Binxing Jiao and Chen Hu and Huacan Wang},
      year={2025},
      eprint={2508.02085},
      archivePrefix={arXiv},
      primaryClass={cs.AI},
      url={https://arxiv.org/abs/2508.02085}, 
}

@article{10.1016/j.neucom.2025.130470,
author = {Liang, Xuechen and Tao, Meiling and Xia, Yinghui and Wang, Jianhui and Li, Kun and Wang, Yijin and He, Yangfan and Yang, Jingsong and Shi, Tianyu and Wang, Yuantao and Zhang, Miao and Wang, Xueqian},
title = {SAGE: Self-evolving Agents with Reflective and Memory-augmented Abilities},
year = {2025},
issue_date = {Sep 2025},
publisher = {Elsevier Science Publishers B. V.},
address = {NLD},
volume = {647},
number = {C},
issn = {0925-2312},
url = {https://doi.org/10.1016/j.neucom.2025.130470},
doi = {10.1016/j.neucom.2025.130470},
journal = {Neurocomput.},
month = sep,
numpages = {12}
}

@inproceedings{liu-etal-2025-contextual,
    title = "Contextual Experience Replay for Self-Improvement of Language Agents",
    author = "Liu, Yitao  and
      Si, Chenglei  and
      Narasimhan, Karthik R  and
      Yao, Shunyu",
    editor = "Che, Wanxiang  and
      Nabende, Joyce  and
      Shutova, Ekaterina  and
      Pilehvar, Mohammad Taher",
    booktitle = "Proceedings of the 63rd Annual Meeting of the Association for Computational Linguistics (Volume 1: Long Papers)",
    month = jul,
    year = "2025",
    address = "Vienna, Austria",
    publisher = "Association for Computational Linguistics",
    url = "https://aclanthology.org/2025.acl-long.694/",
    doi = "10.18653/v1/2025.acl-long.694",
    pages = "14179--14198",
    ISBN = "979-8-89176-251-0"
}

@inproceedings{10.1609/aaai.v38i17.29936,
author = {Zhao, Andrew and Huang, Daniel and Xu, Quentin and Lin, Matthieu and Liu, Yong-Jin and Huang, Gao},
title = {ExpeL: LLM agents are experiential learners},
year = {2024},
isbn = {978-1-57735-887-9},
publisher = {AAAI Press},
url = {https://doi.org/10.1609/aaai.v38i17.29936},
doi = {10.1609/aaai.v38i17.29936},
booktitle = {Proceedings of the Thirty-Eighth AAAI Conference on Artificial Intelligence and Thirty-Sixth Conference on Innovative Applications of Artificial Intelligence and Fourteenth Symposium on Educational Advances in Artificial Intelligence},
articleno = {2188},
numpages = {11},
series = {AAAI'24/IAAI'24/EAAI'24}
}

@misc{ouyang2025reasoningbankscalingagentselfevolving,
      title={ReasoningBank: Scaling Agent Self-Evolving with Reasoning Memory}, 
      author={Siru Ouyang and Jun Yan and I-Hung Hsu and Yanfei Chen and Ke Jiang and Zifeng Wang and Rujun Han and Long T. Le and Samira Daruki and Xiangru Tang and Vishy Tirumalashetty and George Lee and Mahsan Rofouei and Hangfei Lin and Jiawei Han and Chen-Yu Lee and Tomas Pfister},
      year={2025},
      eprint={2509.25140},
      archivePrefix={arXiv},
      primaryClass={cs.AI},
      url={https://arxiv.org/abs/2509.25140}, 
}

@misc{wang2024agentworkflowmemory,
      title={Agent Workflow Memory}, 
      author={Zora Zhiruo Wang and Jiayuan Mao and Daniel Fried and Graham Neubig},
      year={2024},
      eprint={2409.07429},
      archivePrefix={arXiv},
      primaryClass={cs.CL},
      url={https://arxiv.org/abs/2409.07429}, 
}

@misc{cao2025remembermerefineme,
      title={Remember Me, Refine Me: A Dynamic Procedural Memory Framework for Experience-Driven Agent Evolution}, 
      author={Zouying Cao and Jiaji Deng and Li Yu and Weikang Zhou and Zhaoyang Liu and Bolin Ding and Hai Zhao},
      year={2025},
      eprint={2512.10696},
      archivePrefix={arXiv},
      primaryClass={cs.AI},
      url={https://arxiv.org/abs/2512.10696}, 
}

@misc{zhang2025agentlearningearlyexperience,
      title={Agent Learning via Early Experience}, 
      author={Kai Zhang and Xiangchao Chen and Bo Liu and Tianci Xue and Zeyi Liao and Zhihan Liu and Xiyao Wang and Yuting Ning and Zhaorun Chen and Xiaohan Fu and Jian Xie and Yuxuan Sun and Boyu Gou and Qi Qi and Zihang Meng and Jianwei Yang and Ning Zhang and Xian Li and Ashish Shah and Dat Huynh and Hengduo Li and Zi Yang and Sara Cao and Lawrence Jang and Shuyan Zhou and Jiacheng Zhu and Huan Sun and Jason Weston and Yu Su and Yifan Wu},
      year={2025},
      eprint={2510.08558},
      archivePrefix={arXiv},
      primaryClass={cs.AI},
      url={https://arxiv.org/abs/2510.08558}, 
}

@misc{zhai2025agentevolverefficientselfevolvingagent,
      title={AgentEvolver: Towards Efficient Self-Evolving Agent System}, 
      author={Yunpeng Zhai and Shuchang Tao and Cheng Chen and Anni Zou and Ziqian Chen and Qingxu Fu and Shinji Mai and Li Yu and Jiaji Deng and Zouying Cao and Zhaoyang Liu and Bolin Ding and Jingren Zhou},
      year={2025},
      eprint={2511.10395},
      archivePrefix={arXiv},
      primaryClass={cs.LG},
      url={https://arxiv.org/abs/2511.10395}, 
}

@misc{zhang2025agenticcontextengineeringevolving,
      title={Agentic Context Engineering: Evolving Contexts for Self-Improving Language Models}, 
      author={Qizheng Zhang and Changran Hu and Shubhangi Upasani and Boyuan Ma and Fenglu Hong and Vamsidhar Kamanuru and Jay Rainton and Chen Wu and Mengmeng Ji and Hanchen Li and Urmish Thakker and James Zou and Kunle Olukotun},
      year={2025},
      eprint={2510.04618},
      archivePrefix={arXiv},
      primaryClass={cs.LG},
      url={https://arxiv.org/abs/2510.04618}, 
}

@misc{cai2025flexcontinuousagentevolution,
      title={FLEX: Continuous Agent Evolution via Forward Learning from Experience}, 
      author={Zhicheng Cai and Xinyuan Guo and Yu Pei and Jiangtao Feng and Jinsong Su and Jiangjie Chen and Ya-Qin Zhang and Wei-Ying Ma and Mingxuan Wang and Hao Zhou},
      year={2025},
      eprint={2511.06449},
      archivePrefix={arXiv},
      primaryClass={cs.LG},
      url={https://arxiv.org/abs/2511.06449}, 
}

@misc{yang2025learningjobexperiencedrivenselfevolving,
      title={Learning on the Job: An Experience-Driven Self-Evolving Agent for Long-Horizon Tasks}, 
      author={Cheng Yang and Xuemeng Yang and Licheng Wen and Daocheng Fu and Jianbiao Mei and Rong Wu and Pinlong Cai and Yufan Shen and Nianchen Deng and Botian Shi and Yu Qiao and Haifeng Li},
      year={2025},
      eprint={2510.08002},
      archivePrefix={arXiv},
      primaryClass={cs.CL},
      url={https://arxiv.org/abs/2510.08002}, 
}

@misc{xia2025agent0unleashingselfevolvingagents,
      title={Agent0: Unleashing Self-Evolving Agents from Zero Data via Tool-Integrated Reasoning}, 
      author={Peng Xia and Kaide Zeng and Jiaqi Liu and Can Qin and Fang Wu and Yiyang Zhou and Caiming Xiong and Huaxiu Yao},
      year={2025},
      eprint={2511.16043},
      archivePrefix={arXiv},
      primaryClass={cs.LG},
      url={https://arxiv.org/abs/2511.16043}, 
}

@misc{zhang2026darwintodllmdrivenlifelong,
      title={DarwinTOD: LLM Driven Lifelong Self Evolution for Task Oriented Dialog Systems}, 
      author={Shuyu Zhang and Yujie Liu and Xinru Wang and Cheng Zhang and Yanmin Zhu and Bin Li},
      year={2026},
      eprint={2601.07248},
      archivePrefix={arXiv},
      primaryClass={cs.MA},
      url={https://arxiv.org/abs/2601.07248}, 
}

@misc{yue2026drzeroselfevolvingsearch,
      title={Dr. Zero: Self-Evolving Search Agents without Training Data}, 
      author={Zhenrui Yue and Kartikeya Upasani and Xianjun Yang and Suyu Ge and Shaoliang Nie and Yuning Mao and Zhe Liu and Dong Wang},
      year={2026},
      eprint={2601.07055},
      archivePrefix={arXiv},
      primaryClass={cs.AI},
      url={https://arxiv.org/abs/2601.07055}, 
}

@misc{hassell2025learningsupervisionsemanticepisodic,
      title={Learning from Supervision with Semantic and Episodic Memory: A Reflective Approach to Agent Adaptation}, 
      author={Jackson Hassell and Dan Zhang and Hannah Kim and Tom Mitchell and Estevam Hruschka},
      year={2025},
      eprint={2510.19897},
      archivePrefix={arXiv},
      primaryClass={cs.CL},
      url={https://arxiv.org/abs/2510.19897}, 
}

@misc{lu2025searchselfplaypushingfrontier,
      title={Search Self-play: Pushing the Frontier of Agent Capability without Supervision}, 
      author={Hongliang Lu and Yuhang Wen and Pengyu Cheng and Ruijin Ding and Jiaqi Guo and Haotian Xu and Chutian Wang and Haonan Chen and Xiaoxi Jiang and Guanjun Jiang},
      year={2025},
      eprint={2510.18821},
      archivePrefix={arXiv},
      primaryClass={cs.LG},
      url={https://arxiv.org/abs/2510.18821}, 
}

@misc{zhang2026memrlselfevolvingagentsruntime,
      title={MemRL: Self-Evolving Agents via Runtime Reinforcement Learning on Episodic Memory}, 
      author={Shengtao Zhang and Jiaqian Wang and Ruiwen Zhou and Junwei Liao and Yuchen Feng and Weinan Zhang and Ying Wen and Zhiyu Li and Feiyu Xiong and Yutao Qi and Bo Tang and Muning Wen},
      year={2026},
      eprint={2601.03192},
      archivePrefix={arXiv},
      primaryClass={cs.CL},
      url={https://arxiv.org/abs/2601.03192}, 
}

@misc{liu2025trainingenablingselfevolutionagents,
      title={Beyond Training: Enabling Self-Evolution of Agents with MOBIMEM}, 
      author={Zibin Liu and Cheng Zhang and Xi Zhao and Yunfei Feng and Bingyu Bai and Dahu Feng and Erhu Feng and Yubin Xia and Haibo Chen},
      year={2025},
      eprint={2512.15784},
      archivePrefix={arXiv},
      primaryClass={cs.AI},
      url={https://arxiv.org/abs/2512.15784}, 
}

@misc{zheng2024synapsetrajectoryasexemplarpromptingmemory,
      title={Synapse: Trajectory-as-Exemplar Prompting with Memory for Computer Control}, 
      author={Longtao Zheng and Rundong Wang and Xinrun Wang and Bo An},
      year={2024},
      eprint={2306.07863},
      archivePrefix={arXiv},
      primaryClass={cs.AI},
      url={https://arxiv.org/abs/2306.07863}, 
}

@misc{yu2026agenticmemorylearningunified,
      title={Agentic Memory: Learning Unified Long-Term and Short-Term Memory Management for Large Language Model Agents}, 
      author={Yi Yu and Liuyi Yao and Yuexiang Xie and Qingquan Tan and Jiaqi Feng and Yaliang Li and Libing Wu},
      year={2026},
      eprint={2601.01885},
      archivePrefix={arXiv},
      primaryClass={cs.CL},
      url={https://arxiv.org/abs/2601.01885}, 
}

@misc{xu2025amemagenticmemoryllm,
      title={A-MEM: Agentic Memory for LLM Agents}, 
      author={Wujiang Xu and Zujie Liang and Kai Mei and Hang Gao and Juntao Tan and Yongfeng Zhang},
      year={2025},
      eprint={2502.12110},
      archivePrefix={arXiv},
      primaryClass={cs.CL},
      url={https://arxiv.org/abs/2502.12110}, 
}

@misc{westhäußer2025enablingpersonalizedlongterminteractions,
      title={Enabling Personalized Long-term Interactions in LLM-based Agents through Persistent Memory and User Profiles}, 
      author={Rebecca Westhäußer and Wolfgang Minker and Sebatian Zepf},
      year={2025},
      eprint={2510.07925},
      archivePrefix={arXiv},
      primaryClass={cs.AI},
      url={https://arxiv.org/abs/2510.07925}, 
}

@misc{fang2025lightmemlightweightefficientmemoryaugmented,
      title={LightMem: Lightweight and Efficient Memory-Augmented Generation}, 
      author={Jizhan Fang and Xinle Deng and Haoming Xu and Ziyan Jiang and Yuqi Tang and Ziwen Xu and Shumin Deng and Yunzhi Yao and Mengru Wang and Shuofei Qiao and Huajun Chen and Ningyu Zhang},
      year={2025},
      eprint={2510.18866},
      archivePrefix={arXiv},
      primaryClass={cs.CL},
      url={https://arxiv.org/abs/2510.18866}, 
}

@misc{wei2025mlpmemoryretrieverpretrainedmemory,
      title={MLP Memory: A Retriever-Pretrained Memory for Large Language Models}, 
      author={Rubin Wei and Jiaqi Cao and Jiarui Wang and Jushi Kai and Qipeng Guo and Bowen Zhou and Zhouhan Lin},
      year={2025},
      eprint={2508.01832},
      archivePrefix={arXiv},
      primaryClass={cs.CL},
      url={https://arxiv.org/abs/2508.01832}, 
}

@misc{hu2026evermemosselforganizingmemoryoperating,
      title={EverMemOS: A Self-Organizing Memory Operating System for Structured Long-Horizon Reasoning}, 
      author={Chuanrui Hu and Xingze Gao and Zuyi Zhou and Dannong Xu and Yi Bai and Xintong Li and Hui Zhang and Tong Li and Chong Zhang and Lidong Bing and Yafeng Deng},
      year={2026},
      eprint={2601.02163},
      archivePrefix={arXiv},
      primaryClass={cs.AI},
      url={https://arxiv.org/abs/2601.02163}, 
}

@misc{qian2026memobrainexecutivememoryagentic,
      title={MemoBrain: Executive Memory as an Agentic Brain for Reasoning}, 
      author={Hongjin Qian and Zhao Cao and Zheng Liu},
      year={2026},
      eprint={2601.08079},
      archivePrefix={arXiv},
      primaryClass={cs.AI},
      url={https://arxiv.org/abs/2601.08079}, 
}

@misc{zhang2025gmemorytracinghierarchicalmemory,
      title={G-Memory: Tracing Hierarchical Memory for Multi-Agent Systems}, 
      author={Guibin Zhang and Muxin Fu and Guancheng Wan and Miao Yu and Kun Wang and Shuicheng Yan},
      year={2025},
      eprint={2506.07398},
      archivePrefix={arXiv},
      primaryClass={cs.MA},
      url={https://arxiv.org/abs/2506.07398}, 
}

@misc{zhang2025memevolvemetaevolutionagentmemory,
      title={MemEvolve: Meta-Evolution of Agent Memory Systems}, 
      author={Guibin Zhang and Haotian Ren and Chong Zhan and Zhenhong Zhou and Junhao Wang and He Zhu and Wangchunshu Zhou and Shuicheng Yan},
      year={2025},
      eprint={2512.18746},
      archivePrefix={arXiv},
      primaryClass={cs.CL},
      url={https://arxiv.org/abs/2512.18746}, 
}

@misc{ai2025memorybenchbenchmarkmemorycontinual,
      title={MemoryBench: A Benchmark for Memory and Continual Learning in LLM Systems}, 
      author={Qingyao Ai and Yichen Tang and Changyue Wang and Jianming Long and Weihang Su and Yiqun Liu},
      year={2025},
      eprint={2510.17281},
      archivePrefix={arXiv},
      primaryClass={cs.LG},
      url={https://arxiv.org/abs/2510.17281}, 
}

@misc{wei2025evomemorybenchmarkingllmagent,
      title={Evo-Memory: Benchmarking LLM Agent Test-time Learning with Self-Evolving Memory}, 
      author={Tianxin Wei and Noveen Sachdeva and Benjamin Coleman and Zhankui He and Yuanchen Bei and Xuying Ning and Mengting Ai and Yunzhe Li and Jingrui He and Ed H. Chi and Chi Wang and Shuo Chen and Fernando Pereira and Wang-Cheng Kang and Derek Zhiyuan Cheng},
      year={2025},
      eprint={2511.20857},
      archivePrefix={arXiv},
      primaryClass={cs.CL},
      url={https://arxiv.org/abs/2511.20857}, 
}

@misc{xiong2025memorymanagementimpactsllm,
      title={How Memory Management Impacts LLM Agents: An Empirical Study of Experience-Following Behavior}, 
      author={Zidi Xiong and Yuping Lin and Wenya Xie and Pengfei He and Zirui Liu and Jiliang Tang and Himabindu Lakkaraju and Zhen Xiang},
      year={2025},
      eprint={2505.16067},
      archivePrefix={arXiv},
      primaryClass={cs.AI},
      url={https://arxiv.org/abs/2505.16067}, 
}

@article{10.1145/3748302,
author = {Zhang, Zeyu and Dai, Quanyu and Bo, Xiaohe and Ma, Chen and Li, Rui and Chen, Xu and Zhu, Jieming and Dong, Zhenhua and Wen, Ji-Rong},
title = {A Survey on the Memory Mechanism of Large Language Model-based Agents},
year = {2025},
issue_date = {November 2025},
publisher = {Association for Computing Machinery},
address = {New York, NY, USA},
volume = {43},
number = {6},
issn = {1046-8188},
url = {https://doi.org/10.1145/3748302},
doi = {10.1145/3748302},
journal = {ACM Trans. Inf. Syst.},
month = sep,
articleno = {155},
numpages = {47}
}

@misc{yang2025qwen3technicalreport,
      title={Qwen3 Technical Report}, 
      author={An Yang and Anfeng Li and Baosong Yang and Beichen Zhang and Binyuan Hui and Bo Zheng and Bowen Yu and Chang Gao and Chengen Huang and Chenxu Lv and Chujie Zheng and Dayiheng Liu and Fan Zhou and Fei Huang and Feng Hu and Hao Ge and Haoran Wei and Huan Lin and Jialong Tang and Jian Yang and Jianhong Tu and Jianwei Zhang and Jianxin Yang and Jiaxi Yang and Jing Zhou and Jingren Zhou and Junyang Lin and Kai Dang and Keqin Bao and Kexin Yang and Le Yu and Lianghao Deng and Mei Li and Mingfeng Xue and Mingze Li and Pei Zhang and Peng Wang and Qin Zhu and Rui Men and Ruize Gao and Shixuan Liu and Shuang Luo and Tianhao Li and Tianyi Tang and Wenbiao Yin and Xingzhang Ren and Xinyu Wang and Xinyu Zhang and Xuancheng Ren and Yang Fan and Yang Su and Yichang Zhang and Yinger Zhang and Yu Wan and Yuqiong Liu and Zekun Wang and Zeyu Cui and Zhenru Zhang and Zhipeng Zhou and Zihan Qiu},
      year={2025},
      eprint={2505.09388},
      archivePrefix={arXiv},
      primaryClass={cs.CL},
      url={https://arxiv.org/abs/2505.09388}, 
}

@inproceedings{NEURIPS2020_3f5ee243,
 author = {Wang, Wenhui and Wei, Furu and Dong, Li and Bao, Hangbo and Yang, Nan and Zhou, Ming},
 booktitle = {Advances in Neural Information Processing Systems},
 editor = {H. Larochelle and M. Ranzato and R. Hadsell and M.F. Balcan and H. Lin},
 pages = {5776--5788},
 publisher = {Curran Associates, Inc.},
 title = {MiniLM: Deep Self-Attention Distillation for Task-Agnostic Compression of Pre-Trained Transformers},
 url = {https://proceedings.neurips.cc/paper_files/paper/2020/file/3f5ee243547dee91fbd053c1c4a845aa-Paper.pdf},
 volume = {33},
 year = {2020}
}

@inproceedings{NEURIPS2023_43e9d647,
 author = {Liu, Jiawei and Xia, Chunqiu Steven and Wang, Yuyao and ZHANG, LINGMING},
 booktitle = {Advances in Neural Information Processing Systems},
 editor = {A. Oh and T. Naumann and A. Globerson and K. Saenko and M. Hardt and S. Levine},
 pages = {21558--21572},
 publisher = {Curran Associates, Inc.},
 title = {Is Your Code Generated by ChatGPT Really Correct? Rigorous Evaluation of Large Language Models for Code Generation},
 url = {https://proceedings.neurips.cc/paper_files/paper/2023/file/43e9d647ccd3e4b7b5baab53f0368686-Paper-Conference.pdf},
 volume = {36},
 year = {2023}
}

\clearpage
\appendix
\section{Algorithmic Details}
Algorithm~\ref{alg:hint_construction} details the procedure for constructing structured experience hints from a ranked set of retrieved cases under a fixed budget. This design ensures that limited prompt capacity is allocated preferentially to experience instances that convey concrete corrective knowledge, while retaining flexibility to incorporate additional relevant cases when budget allows.

\begin{algorithm}[b]
\caption{Hint Construction from Ranked Cases.}
\label{alg:hint_construction}
\begin{algorithmic}[1]
\Require Ranked cases $\mathcal{C}_{\mathrm{rank}}$, correction relation $\mathcal{E}_{\textit{fix}}$ (fixed\_by), hint budget $M$
\Ensure Structured hint set $\mathcal{H}$

\State $\mathcal{S} \leftarrow [\,]$ \Comment{selected cases for hint construction}
\State $\mathcal{W}_{\text{fix}} \leftarrow [\,]$ \Comment{warning cases with a fixed\_by relation}

\ForAll{$c \in \mathcal{C}_{\mathrm{rank}}$}
    \If{$c$ is a warning case \textbf{and} $\exists (c \rightarrow c^{+}) \in \mathcal{E}_{\textit{fix}}$}
        \State append $c$ to $\mathcal{W}_{\text{fix}}$
        \If{$|\mathcal{W}_{\text{fix}}|$ reaches the configured limit}
            \State \textbf{break}
        \EndIf
    \EndIf
\EndFor

\ForAll{$c \in \mathcal{W}_{\text{fix}}$}
    \State add $c$ to $\mathcal{S}$
    \If{$|\mathcal{S}| \ge M$}
        \State \Return $\mathrm{FormatHint}(\mathcal{S}, \mathcal{E}_{\textit{fix}})$
    \EndIf
\EndFor

\If{including corrected counterparts}
    \ForAll{$c \in \mathcal{W}_{\text{fix}}$}
        \State $c^{+} \leftarrow$ the unique case such that $(c \rightarrow c^{+}) \in \mathcal{E}_{\textit{fix}}$
        \If{$c^{+}$ exists}
            \State add $c^{+}$ to $\mathcal{S}$
        \EndIf
        \If{$|\mathcal{S}| \ge M$}
            \State \Return $\mathrm{FormatHint}(\mathcal{S}, \mathcal{E}_{\textit{fix}})$
        \EndIf
    \EndFor
\EndIf

\ForAll{$c \in \mathcal{C}_{\mathrm{rank}}$}
    \State add $c$ to $\mathcal{S}$ \Comment{deduplicate while adding}
    \If{$|\mathcal{S}| \ge M$}
        \State \textbf{break}
    \EndIf
\EndFor

\State $\mathcal{H} \leftarrow \mathrm{FormatHint}(\mathcal{S}, \mathcal{E}_{\textit{fix}})$
\State \Return $\mathcal{H}$
\end{algorithmic}
\end{algorithm}

\section{Datasets and Models}
\label{data}
\subsection{Datasets}

\textbf{HumanEval} is a standard benchmark for evaluating functional program synthesis by large language models. It consists of 164 hand-written Python programming tasks, each defined by a natural language description, a function signature, and a set of hidden unit tests used for automatic evaluation. The benchmark focuses on semantic correctness rather than surface-level similarity, as generated programs must pass all test cases to be considered correct. HumanEval is particularly suitable for studying experience reuse, since many tasks share recurring programming patterns and failure modes, allowing prior successes and corrections to inform subsequent code generation attempts.

\textbf{EvalPlus} is an enhanced evaluation suite built upon HumanEval, designed to provide stricter and more comprehensive correctness verification. It augments each original problem with additional test cases, including adversarial and edge-case inputs, to reduce false positives caused by insufficient test coverage. In this work, EvalPlus is used to assess whether the benefits of experience reuse persist under more rigorous functional validation, ensuring that observed improvements reflect genuine semantic correctness rather than overfitting to limited tests.

\textbf{MuSiQue} (Multi-hop Sequential Question Answering) is a benchmark designed to evaluate compositional multi-hop reasoning over text. Each question is constructed to require a specific reasoning chain involving multiple supporting facts, often combining entity linking, relational reasoning, and temporal inference. By explicitly controlling reasoning depth and discouraging shortcut solutions, MuSiQue provides a challenging testbed for analyzing whether accumulated experience can be abstracted into reusable reasoning patterns across structurally similar multi-hop questions.

\textbf{HotpotQA} is a large-scale multi-hop question answering dataset built from Wikipedia articles, containing over 100k question--answer pairs. A key characteristic of HotpotQA is the annotation of sentence-level supporting facts, enabling supervision and evaluation of explainable reasoning. The dataset includes both bridge-type questions, which require identifying intermediate entities, and comparison-type questions that integrate information across documents. In our experiments, HotpotQA serves to evaluate whether experience graphs can capture and reuse structural reasoning signals, such as missing hops or incorrect entity associations, in complex multi-document inference tasks.

\subsection{Models}

All models used in this work are accessed exclusively through APIs and model selection is guided by dataset difficulty, task specialization, and a balanced consideration of open-weight versus proprietary models.

\textbf{Qwen3-1.7B} is a small-scale open-weight model in the Qwen3 family, designed for efficient inference under strict computational constraints. We include Qwen3-1.7B to examine whether structured experience reuse remains effective at very limited model capacity, providing a lower-bound assessment of EXG’s robustness.

\textbf{Qwen3-8B} is a compact open-weight model in the Qwen3 family, with approximately 8B parameters, designed to provide strong general reasoning capabilities under constrained computational budgets. In our experiments, Qwen3-8B is primarily used to assess the robustness of experience reuse mechanisms under limited model capacity.

\textbf{Qwen3-14B} is a mid-to-large scale open-weight model in the Qwen3 family, positioned between compact models such as Qwen3-8B and larger-capacity models like Qwen-Max. It offers stronger reasoning and representation capacity than smaller Qwen3 variants while remaining more computationally efficient than frontier-scale models. In our experiments, Qwen3-14B is used to evaluate the effectiveness of structured experience reuse at an intermediate model scale, serving as a bridge between lightweight and high-capacity backbones.

\textbf{Qwen3-Coder-Flash} is a lightweight, code-oriented model optimized for fast inference and program synthesis tasks. It has an approximate parameter scale of \(\sim 30\)B and is designed to efficiently handle function-level code generation with low latency. In our experiments, Qwen3-Coder-Flash is mainly used on HumanEval and EvalPlus, where rapid iteration and clear execution feedback are essential.

\textbf{Qwen-Plus} is a mid-sized general-purpose language model that offers a balance between reasoning capability and computational efficiency, making it suitable for reasoning-intensive tasks. In our experiments, Qwen-Plus is used primarily for question answering benchmarks, including MuSiQue and HotpotQA.

\textbf{Qwen-Max} is a large-capacity model in the Qwen3 family and it provides stronger reasoning and comprehension abilities, particularly for complex language understanding and multi-hop inference. In our experiments, Qwen-Max is used for multi-hop question answering tasks, including MuSiQue and HotpotQA.

\section{Additional Experimental Analysis}

\begin{figure}[b]
    \centering
    \includegraphics[width=0.9\columnwidth]{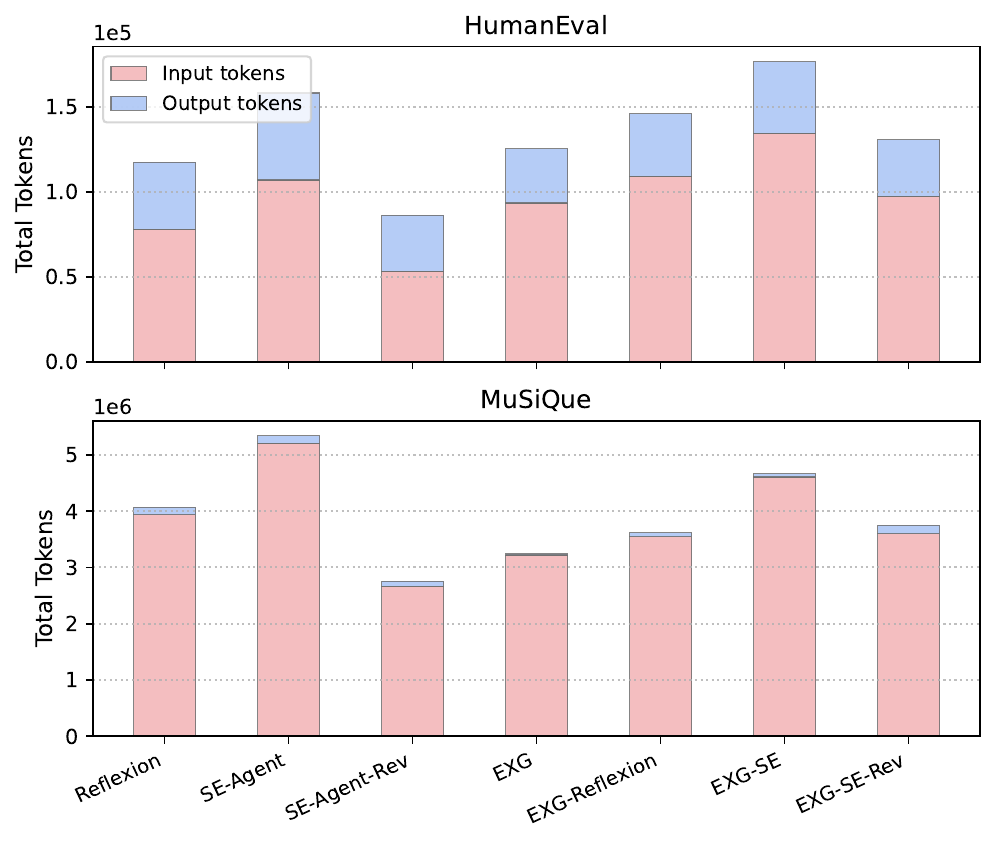}
    \caption{Token usage breakdown on HumanEval and MuSiQue. Each bar shows the total number of tokens consumed, with input tokens stacked below output tokens.}
    \label{fig:token-breakdown}
\end{figure}

\subsection{Online Token Usage}
To better understand how structured experience reuse affects computational cost, we analyze token consumption across different methods and datasets, focusing on how input and output tokens are redistributed under a fixed interaction budget.

On HumanEval, baseline methods without structured experience reuse exhibit substantial token overhead due to repeated full-code generation. Reflexion consumes 117{,}378 total tokens across all tasks, whereas EXG consumes 125{,}304 tokens, corresponding to a modest increase of 6.8\% in total token usage. This increase is driven by a deliberate rise in input tokens (+20.0\%), reflecting the injection of experience hints, while output tokens are reduced by 19.3\%. Compared to SE-Agent-Lite, which incurs 158{,}125 total tokens, EXG reduces total token consumption by 20.8\%, with a 37.8\% reduction in output tokens. These results indicate that, on HumanEval, EXG substantially reduces redundant generations, and that its slightly higher prompt cost is more than offset when compared to retry- and revision-based baselines.

The impact of experience reuse is significantly amplified on MuSiQue, a long-context multi-hop reasoning benchmark. Reflexion consumes 4.07M total tokens, while EXG reduces this to 3.24M tokens, yielding a 20.4\% reduction in total token usage. This improvement is almost entirely attributable to a dramatic reduction in output tokens (84.0\%), despite an 18.3\% increase in input tokens. Relative to SE-Agent-Lite, which exceeds 5.34M total tokens, EXG achieves an even larger reduction of 39.4\%. These figures highlight that, in long-horizon reasoning tasks, avoiding repeated reasoning chains yields substantial absolute token savings.

Across both datasets, EXG consistently shifts computation from output-heavy repeated generation to input-side structured guidance. While this shift leads to moderate increases in prompt length, it results in large and consistent reductions in output tokens, ranging from 19.3\% on HumanEval to over 80\% on MuSiQue. The magnitude of output-token reduction scales with task complexity, demonstrating that the efficiency gains of EXG arise from reduced redundant exploration rather than dataset-specific effects.

Taken together, these results show that EXG improves token efficiency not by compressing individual prompts or enforcing shorter outputs, but by altering the interaction dynamics of self-evolving agents. By leveraging accumulated experience to guide generation earlier, EXG reduces repeated attempts and enables experience to compound over time, yielding superior accuracy--efficiency trade-offs, especially in long-context reasoning settings.

\subsection{Online Learning Curve}

On HumanEval, all methods start from a similar performance level in the early stage of deployment. After the first 20 tasks, cumulative Pass@1 for Reflexion, SE-Agent, and EXG-based methods all fall within a narrow range of approximately 50--55\%. As more tasks are observed, baseline methods exhibit only marginal improvement: after 60 tasks, Reflexion remains around 56--58\% and shows little further increase. In contrast, EXG-based methods demonstrate a sustained upward trend. By 60 tasks, EXG-based methods reach approximately 68--70\%, corresponding to an absolute improvement of about 14--15 percentage points over its early-stage performance. This gap continues to widen as deployment proceeds. By the end of the task sequence, EXG-based methods attain a cumulative Pass@1 of approximately 80\%, while baseline methods plateau near 55--58\%. Overall, EXG improves Pass@1 by roughly 25 percentage points from early to late stages, whereas baselines improve by less than 5 points.

A similar but more pronounced pattern is observed for Pass@2. At the early stage (20 tasks), all methods achieve comparable Pass@2 values of approximately 72--76\%. Baseline methods quickly saturate: Reflexion stabilizes near 75--78\% after 40--60 tasks, with negligible improvement thereafter. In contrast, EXG-based methods continue to improve as more tasks are seen. By around 60 tasks, EXG-based methods reach approximately 85\%, already exceeding baseline performance by about 7--10 percentage points. By the end of the task sequence, EXG-based methods achieve a cumulative Pass@2 close to 90\%, compared to 75--78\% for baseline methods. This corresponds to an absolute late-stage improvement of roughly 12--15 percentage points over Reflexion, indicating that experience reuse compounds the effectiveness of limited retries over time.

\begin{figure}[t]
    \centering
    \includegraphics[width=\columnwidth]{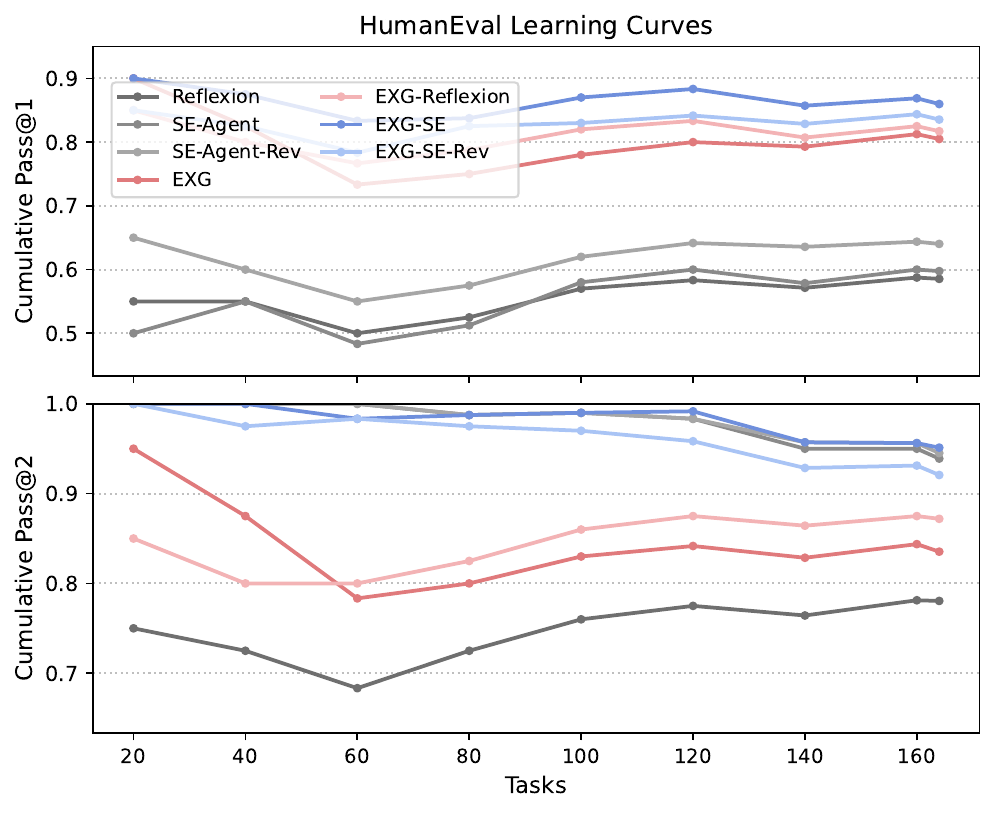}
    \caption{Learning curves on HumanEval.}
    \label{fig:learning-curve-humaneval}
\end{figure}

\begin{figure}[b]
    \centering
    \includegraphics[width=\columnwidth]{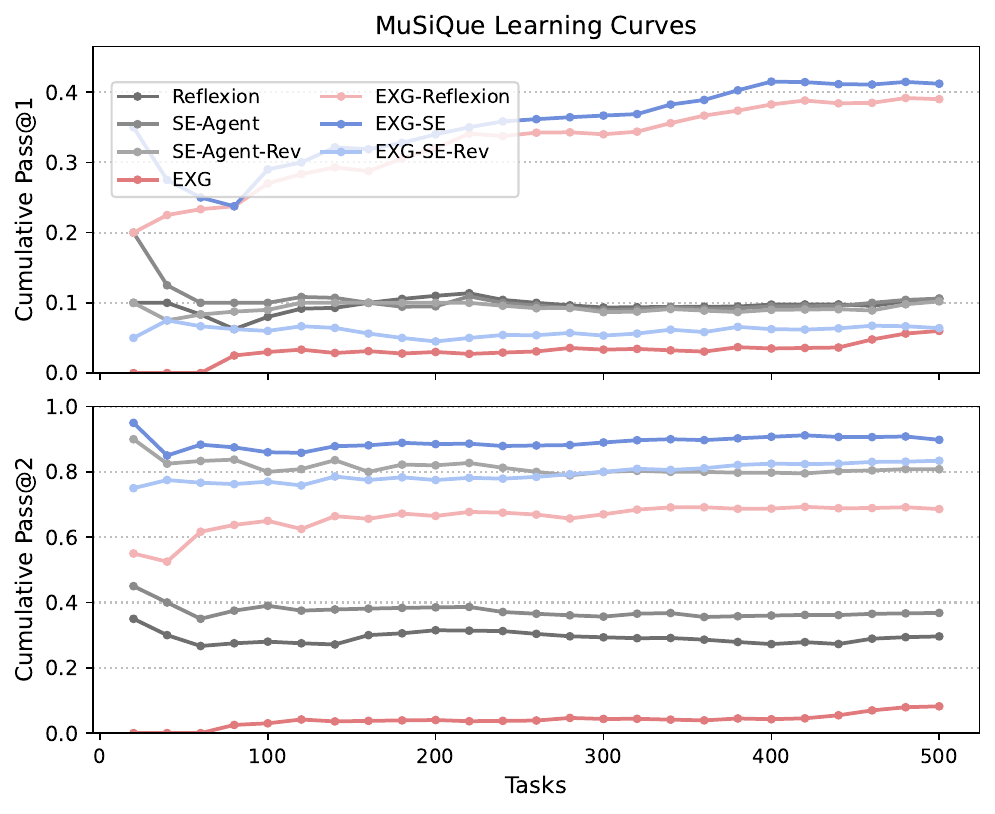}
    \caption{Learning curves on MuSiQue.}
    \label{fig:learning-curve-musique}
\end{figure}

MuSiQue presents a more challenging setting in which early performance remains low for all methods, reflecting the difficulty of long-context, multi-hop reasoning. During the initial phase, cumulative Pass@1 for both baseline methods and EXG-based methods remains clustered in a narrow band of roughly 30--35\%, suggesting that little reusable structure is available at the beginning. As task exposure increases, baseline methods exhibit limited progression: even after around 60 tasks, their Pass@1 rises only marginally to approximately 36--38\%, after which further gains largely diminish. In contrast, EXG-based methods collectively display a qualitatively different trajectory. The core EXG-based method shows a consistent upward trend as experience accumulates, reaching approximately 45--47\% Pass@1 by 60 tasks---an absolute gain of about 10--12 percentage points relative to its early-stage performance. Variants that combine EXG with additional self-evolution mechanisms (e.g., reflection or revision) follow a similar but slightly more variable pattern, indicating that the primary driver of improvement is the experience graph itself rather than auxiliary operators. By the end of the task sequence, EXG-based methods approach 50\% Pass@1, while baseline methods remain below 40\%, yielding a late-stage advantage of approximately 12--15 percentage points.

Allowing a second attempt further accentuates the separation between baseline and EXG-based methods. In the early stage, Pass@2 values for all methods lie in a comparable range of approximately 45--50\%, indicating that limited retries alone do not overcome the inherent difficulty of the task. Baseline methods quickly saturate: Reflexion stabilizes around 50--52\% after 40--60 tasks, with little subsequent improvement.
By contrast, EXG-based methods continue to benefit from accumulated experience throughout deployment. The core EXG method reaches approximately 60--62\% Pass@2 by 60 tasks, exceeding baseline performance by 8--10 percentage points at the same stage. EXG-based variants follow closely, with minor fluctuations attributable to their additional control logic. By the end of deployment, EXG-based methods achieve cumulative Pass@2 values near 65\%, compared to 50--52\% for baseline methods, corresponding to a late-stage improvement of approximately 13--15 percentage points. These results indicate that structured experience reuse not only improves first-attempt reasoning, but also increasingly enhances the effectiveness of limited retries in long-horizon multi-hop settings.

\subsection{Graph Statistics}

\begin{table*}[t]
\centering
\caption{Experience graph statistics averaged over task-level graph snapshots.}
\label{tab:graph-stats}
\setlength{\tabcolsep}{7pt}
\renewcommand{\arraystretch}{1}
\begin{tabular}{lcccccc}
\toprule
& \multicolumn{3}{c}{\textbf{HumanEval}} & \multicolumn{3}{c}{\textbf{MuSiQue}} \\
\cmidrule(lr){2-4} \cmidrule(lr){5-7}
\textbf{Method}
& \textbf{Cases}
& \textbf{\emph{similar\_to}}
& \textbf{\emph{fixed\_by}}
& \textbf{Cases}
& \textbf{\emph{similar\_to}}
& \textbf{\emph{fixed\_by}} \\
\midrule
EXG
& 99.9
& 905
& 3.6
& 491.8
& 5{,}021
& 2.9 \\

EXG-Reflexion
& 98.1
& 880
& 3.4
& 412.4
& 3{,}937
& 81.3 \\

EXG-Revision
& 96.5
& 876
& 10.4
& 486.0
& 4{,}692
& 187.1 \\

EXG-SE
& 93.9
& 843
& 9.5
& 406.4
& 3{,}926
& 129.7 \\
\bottomrule
\end{tabular}
\end{table*}

Table~\ref{tab:graph-stats} provides quantitative evidence on how the experience graph evolves across datasets. On both HumanEval and MuSiQue, the number of case nodes closely tracks the number of processed tasks, with an average of around 95--100 cases on HumanEval and around 400--500 cases on MuSiQue across EXG-based methods. This confirms that graph growth is approximately linear in the number of tasks rather than in the number of attempts, despite cases being defined at the attempt level. As a result, the experience graph remains compact and task-aligned throughout deployment, avoiding uncontrolled expansion due to repeated retries.

Beyond graph size, the two datasets exhibit a strikingly similar local connectivity pattern. After collapsing symmetric similarity relations into undirected edges, each case is connected to around 18--20 similar cases on average, even though the absolute number of cases differs by nearly a factor of five. This consistency indicates that EXG induces a stable similarity backbone that is largely independent of dataset scale or domain. Such dense local neighborhoods allow experience to propagate across related cases, providing a structural explanation for the steadily improving learning curves observed in both code generation and multi-hop reasoning tasks.

In contrast, correction relations exhibit substantial variation across datasets and EXG-based methods. On HumanEval, the number of \emph{fixed\_by} edges remains consistently low, ranging from around 3 to 10, corresponding to roughly 3--11\% of cases participating in explicit correction relations. On MuSiQue, however, the number of \emph{fixed\_by} edges varies much more widely, from around 3 for the core EXG method to over 180 when combined with reflection- or revision-based strategies. This increase is particularly pronounced for reflection-based architectures, as explicit reflection after a failed attempt makes it easier to align a \emph{warning} case with its corrected \emph{golden} counterpart, thereby facilitating the construction of \emph{fixed\_by} relations. As a result, while similarity relations form a stable backbone for experience propagation, correction relations are more sensitive to auxiliary self-evolution mechanisms. Crucially, even when reflection substantially increases the number of \emph{fixed\_by} edges, these relations remain sparse relative to overall graph size, indicating that EXG-based methods rely on amplifying a limited number of well-aligned, high-impact fixes rather than on frequent trial-and-error corrections.

\end{document}